\documentclass[journal]{IEEEtran}
\usepackage{graphicx}
\usepackage{algorithmic}
\usepackage{algorithm}
\usepackage{textcomp}
\usepackage{booktabs}
\usepackage{verbatim}
\usepackage{multirow}
\usepackage{lineno}
\usepackage{amsmath,amsfonts,amssymb}
\usepackage{setspace}
 \usepackage{cite}

\ifCLASSINFOpdf
\else
\fi
\begin{document}
%
\title{Learning Efficient  Representations for Enhanced Object Detection on Large-scene SAR Images}

\author{Siyan~Li,
        Yue~Xiao,
        Yuhang~Zhang,
        Lei~Chu,~\IEEEmembership{Member,~IEEE}
        and~Robert~C.~Qiu,~\IEEEmembership{Fellow,~IEEE}
\thanks{Siyan Li, Yue Xiao, Yuhang Zhang, Lei Chu and Robert C. Qiu were with the center for Big Data and Artificial Intelligence, School of Electronics, Information and Electrical Engineering,
Shanghai Jiaotong University, Shanghai, 200240 China, e-mail: siyan\_li@sjtu.edu.cn}
\thanks{Manuscript received April 19, 2005; revised August 26, 2015.}}

\markboth{Journal of \LaTeX\ Class Files,~Vol.~14, No.~8, August~2015}%
{Shell \MakeLowercase{\textit{et al.}}: Bare Demo of IEEEtran.cls for IEEE Journals}

\maketitle

\begin{abstract}
It is a challenging problem to detect and recognize targets on complex large-scene Synthetic Aperture Radar (SAR) images. Recently developed deep learning algorithms can automatically learn the intrinsic features of SAR images, but still have much room for improvement on large-scene SAR images with limited data. In this paper, based on learning representations and multi-scale features of SAR images, we propose an efficient and robust deep learning based target detection method. Especially, by leveraging the effectiveness of adversarial autoencoder (AAE) which influences the distribution of the investigated data explicitly, the raw SAR dataset is augmented into an enhanced version with a large quantity and diversity. Besides, an auto-labeling scheme is proposed to improve labeling efficiency.
Finally, with jointly training small target chips and large-scene images, an integrated YOLO network combining non-maximum suppression on sub-images is used to realize multiple targets detection of high resolution images. The numerical experimental results on the MSTAR dataset show that our method can realize target detection and recognition on large-scene images accurately and efficiently. The superior anti-noise performance is also confirmed by experiments.
\end{abstract}


%
\IEEEpeerreviewmaketitle

\section{Introduction}
%
%
%
%
\label{1}  
\IEEEPARstart{T}{arget} recognition on SAR images has been under research for many years \cite{wang2016robust,liu2018sar,zhang2018fast,zhao2018adaptive,8533426} due to its various applications in military and homeland security, such as friend and foe identification, battlefield surveillance, environmental monitoring, disaster relief, etc. And it can operate under all-weather and all-time conditions while producing high resolution images with a long standoff capability. Therefore, the SAR image interpretation is of critical importance and the development of automatic target recognition (ATR) system is practical and necessary.

The typical Synthetic Aperture Radar automatic target recognition (SAR-ATR) system can be divided into three parts: target detection, target discrimination and target classification \cite{dudgeon1993overview}. In the first part, a constant false alarm rate (CFAR) detector is used to extract potential targets from SAR images. These potential targets not only consist of true targets such as  armored vehicles, rocket launcher and tanks, but also some background clutters such as trees, buildings and rivers. To reduce false alarm rate, the second discrimination part is designed to train a two-class (target and background) model into capturing the true targets by feature extraction. Finally, the third classification part helps decide which category the target belongs to. However, the traditional SAR-ATR system has several disadvantages \cite{morgan2015deep}. Firstly, it relies heavily on handcrafted features, needs large computational space and has poor robustness. Besides, the accuracy will degrade significantly while any of these three stages is not well designed. Lastly, when it comes to both localizing and classifying the multiple targets in the complex background, it is neither effective nor efficient.

To solve this problem, a novel Moving and Stationary Target Acquisition and Recognition (MSTAR) system was developed by the Air Force Research Laboratory and the Defense Advanced Research Projects Agency (AFRL/DARPA) \cite{article}. This dataset contains not only small target chips that are abstracted from the collected data but also simple and complex large-scene backgrounds since it is costly to directly acquire large-scene SAR images with targets. Based on this dataset, a lot of experiments have been conducted which can be summarized into two aspects: classification on small target chips and detection on synthesized large-scene SAR images \cite{cui2019d}. 

With the emergence of deep learning methods, neural networks have been gradually applied to those two aspects owing to its superior performance on SAR image processing \cite{wen2018survey, 1202937}. Different from the traditional feature extraction methods which need to design the algorithms manually, neural networks are capable of capturing the inherent feature of the input images. As to the first aspect which is to classify the targets on small target chips, the most commonly used deep learning architecture CNN model \cite{inproceedings} is adopted to conduct ten-class classification on MSTAR target chips, which verifies the validity of deep neural network in the field of SAR target recognition. However, the sample number of each type is limited, thus the experimental results lack some commonality. To tackle with the problem of limited training data, domain-specific data augmentation operations combined with CNN \cite{ding2016convolutional} provides a new way to deal with the problem of translation of target, randomness of speckle noise and lack of pose images together. Since a large amount of data is necessary to train a CNN model, another way to deal with the problem of the limited data is to train a ConvNets model with fewer degrees of freedom by only using a sparsely connected convolution architecture \cite{chen2016target} and in the meanwhile randomly sampling relatively smaller patches from the original SAR images to expand the training set. 

As in the above methods, the commonly used data augmentation approaches are horizontal flipping, randomly cropping, rotation, translation or randomly sampling, which means we need to manually control the variety of the additional images by randomly deciding on how many and which ways we are to use. Recently, a newly appeared Generative Adversarial Nets (GAN) proposed by Goodfellow \cite{goodfellow2014generative} is employed to produce more labeled SAR data \cite{he2019parallel}. Though thousands of data can be generated conveniently, not all of them are helpful for classification, so a certain number of generated samples should be carefully selected and it is difficult to find an objective standard to evaluate the quality of the generated images. To avoid the dilemma, another way to make full use of GAN is to train a super-resolution generative adversarial network (SRGAN) \cite{ShiAutomatic} directly to enhance the original images and improve the visual resolution and feature characterization ability of targets in the SAR images. These two methods verify the effective application of the adversarial networks in the SAR image recognition area.

However, GAN-based models have several disadvantages: first, they operate on observation space, which means a large number of parameters are needed during the training process, making it hard to converge; second, due to the high-noise characteristic of SAR dataset, the latent space is more able to capture the main feature of the target in the image which excludes the disturbance of the background. For the sake of solving these two problems, we use a new generative model called Adversarial Autoencoders (AAE) \cite{makhzani2015adversarial}. Different from GANs, AAE blazes a new trail by making the most of latent space. It absorbs the idea of autoencoder \cite{li2017prediction, xiao2020deep} and attempts to push the latent vector close to the distribution of the specific input sample. In this way, AAE is much easier to converge and consumes less space, and our experiment further shows that it also reaches higher quality on generated SAR images. Therefore, in this paper, the AAE network is used to realize data augmentation, and experiments are conducted for improvement on complex large-scene SAR images detection.

So far the above SAR-ATR algorithms are nearly all constructed on CNN framework and the main goal is to classify the targets after the corresponding small chips are abstracted from real large-scene images. In real conditions, however, the targets are randomly scattered into different areas in a real large-scene image with high resolution, and the complex background including trees, buildings, rivers and so on makes it rather hard to accurately detect and recognize them in real-time. Therefore it is under critical research for detecting and recognizing targets on complex large-scene SAR images. Two kinds of algorithms are widely used: two-stage ones, such as R-CNN series \cite{girshick2014rich,girshick2015fast,ren2015faster} and one-stage ones, e.g., SSD \cite{liu2016ssd} and YOLO series \cite{redmon2016you, redmon2017yolo9000,redmon2018yolov3}. The two-stage method Faster R-CNN generally reaches higher accuracy than one-stage methods SSD and YOLOv3 but is time-consuming, too computationally intensive for embedded systems and not suitable for real-time applications. Modified Faster R-CNN models and single shot multibox detector (SSD) are conducted to address SAR-ATR \cite{dong2019end}. It has been shown that MobileNet-SSD and SSD-Inception though have lower accuracy, perform hundreds of times faster than Faster R-CNNs. The work of \citenum{redmon2018yolov3} proposed an improved YOLO network which is known as YOLOv3. This network derives from the older version of YOLO with unique features such as bounding box prediction, class prediction, predictions across scales, feature extractor and training method. The experiment shows that it is three times faster than SSD on COCOs while reaching close detection accuracy. So far, the YOLOv3 network has proved its superiority in many fields such as novel landmark localization \cite{huang2015densebox}, 3-D human detection\cite{Tian2018Robust}, and thermal imaging  \cite{ivavsic2019human}. Following aforementioned state-of-art works in the literature, in this paper, we adopt the YOLOv3 as the backbone for realizing effective and efficient SAR-ATR. 

When it comes to detecting multiple objects in a large complex SAR background \cite{chang2010change, xu2019retrieval}, a fast sliding method can be used to segment the scene image into sub-images and then detection network is applied to locate the targets. The process of target segmentation and synthesis is of rather importance since it is costly to directly gain the large-scene SAR images with multiple targets inside, therefore this process plays a critical part in the final detection and recognition result.

In this paper, we propose a deep learning framework for detection and recognition on complex large-scene SAR images. Before training the network, AAE is firstly adopted to realize the data augmentation of small SAR chips. Such an operation is simple but useful for the extraction of key feature  and enhancing the variety of generated images. In addition, instead of manual labeling, an automatic labeling method is then proposed to mark the targets. Due to the limited number of complex large-scene SAR images, we fully take advantage of small chips and then propose a target segmentation and synthesis method to establish a complex large-scene SAR database for study. After establishing the database, a fast sliding method on large-scene images is proposed to avoid obtaining abundant slices without targets or with incomplete targets. When training the YOLOv3 network, we pretrain the weights of the proposed deep learning method using the well-known COCO dataset by leveraging the advantages of the transfer learning \cite{wang2018sar}. At the training stage, the expanded small target chips and large-scene images after fast sliding are simultaneously fed into the network. Finally, non-maximum suppression on sub-images is conducted to obtain the unique bounding box for each target. The results show that our method exhibits superior accuracy on complex large-scene images and also demonstrates great real-time performance. Furthermore, numerical simulations demonstrate that the proposed method can accurately detect and recognize the targets with high anti-noise performance. 

The remainder of this paper is organized as follows. Section \ref{2} elaborates a target detection and recognition framework for complex large-scene SAR images. In Sec. \ref{3}, we verify the effectiveness and efficiency of our proposed approach on a variety of experiments using the MSTAR dataset. The analysis and conclusions are drawn in Sec. \ref{4}.

\section{The ATR Framework}\label{2}
In this section, we will introduce our target detection and recognition framework on complex large-scene SAR images. Since we need to obtain small target chips for joint training, we will first introduce how to expand SAR target chips and conduct automatic labeling in Sec. \ref{2-1}. Then Sec. \ref{2-2} gives a further description of how to establish our large-scene SAR database, and use YOLOv3 for detection.
\subsection{Process on Small SAR Target Chips}\label{2-1}

\begin{figure}[]
	\centering
	\includegraphics[width=3.5in]{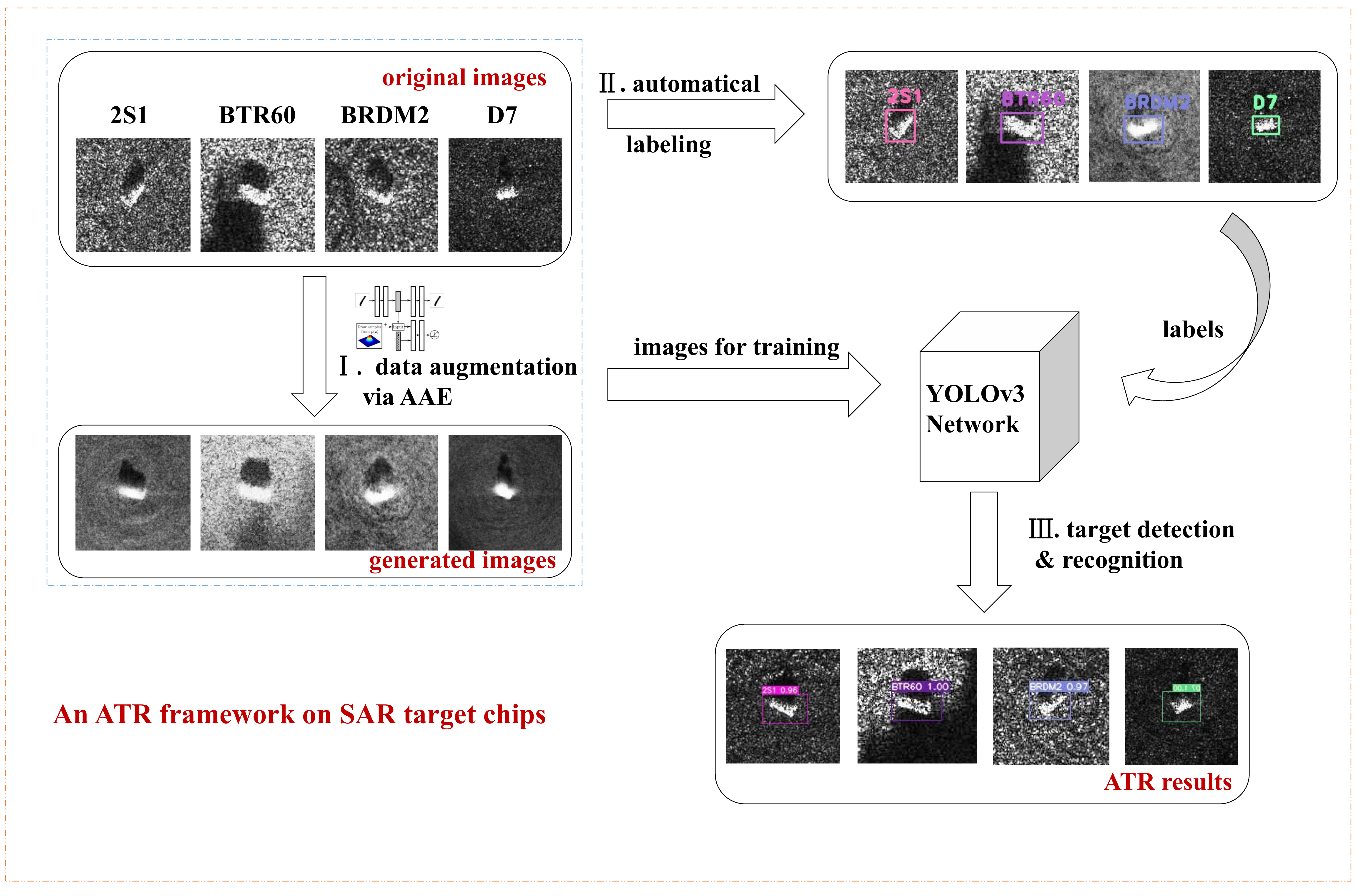}
	\caption{An ATR framework or SAR target chips. \uppercase\expandafter{\romannumeral1}. Data augmentation method AAE is used to expand the training set; \uppercase\expandafter{\romannumeral2}. The training samples are then automatically labeled; \uppercase\expandafter{\romannumeral3}. The training images and labels are sent to the YOLOv3 network to complete target detection and recognition.}
	\label{chips}
\end{figure}

The proposed ATR model on SAR target chips is shown in Fig. \ref{chips}. It is composed of three parts: data augmentation by AAE, automatic labeling, target detection and recognition. The last part is realized by YOLOv3 after automatically labeling these targets, which means we can not only detect the target but also recognize it with limited samples and without manual labeling. These expanded labeled small chips are fed into the network with large-scene images to enhance the detection accuracy on complex background. 

In Fig. \ref{chips}, we use MSTAR four-target dataset, including 2S1, BTR60, BRDM2 and D7 as an example to illustrate the AAE augmentation method and automatic labeling.

\subsubsection{Data augmentation}

One of the key points in SAR image recognition is that SAR images suffer from the speckle noise due to the characteristic of the imaging system. And SAR images for training a robust ATR system is insufficient. For instance, in MSTAR four-target dataset, there are only 1152 images for training, which may lead to an overfitting problem and reduce the generalization effect.

To solve the problem of insufficient training samples, data augmentation is necessary. The classic methods of data augmentation are mostly operating on the original images through flipping, cropping, zooming, etc. This may result in data redundancy and therefore can not obviously enhance the variety of image characteristics. 

\begin{figure}[]
	\centering
	\includegraphics[width=3in]{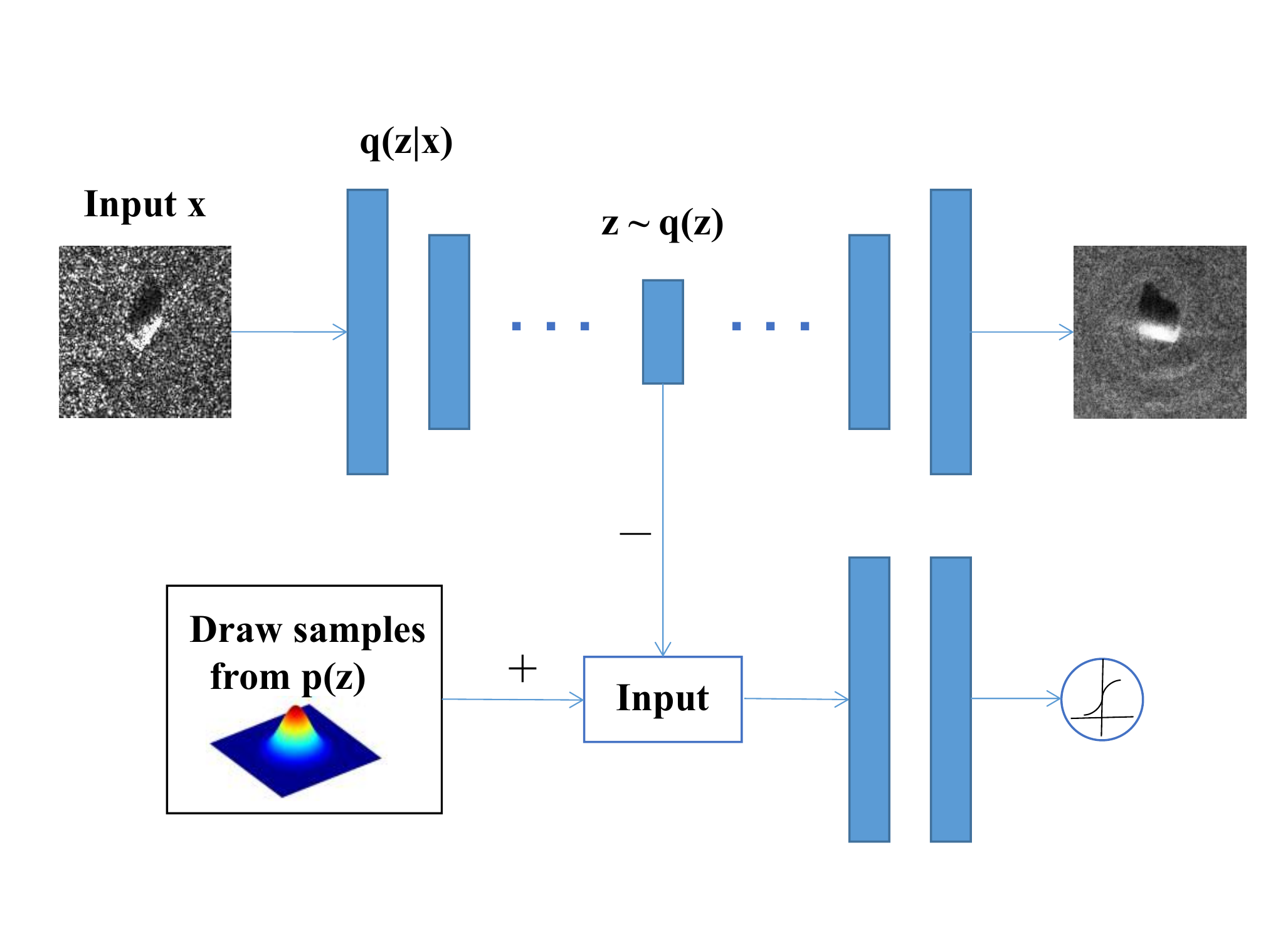}
	\caption{The architecture of an adversarial autoencoder. The top row is a standard autoencoder that reconstructs an image $x$ from a latent code $z$. The bottom row is a network trained to discriminate whether the sample is from a prior distribution or from the latent vector $z$. }
	\label{AAE}
\end{figure}

The adversarial autoencoder (AAE) is a combination of autoencoder and GAN, and it achieves competitive performance on generating SAR target chips. As is shown in Fig. \ref{AAE}, the top row is a standard autoencoder that reconstructs an image $x$ from a latent code $z$.
\begin{equation}
\label{eqn_example}
q(z) = \int_x {q(z|x){p_d}(x)dx} 
\end{equation}

The goal of the adversarial autoencoder is to match the aggregated posterior $q(z)$ to $p(z)$, which is an arbitrary prior (e.g. Gaussian distribution). The encoder of the autoencoder ${q(z|x)}$ acts as the generator of the adversarial network, attempting to fool the discriminative adversarial network into recognizing the hidden code $q(z)$ as the prior distribution $p(z)$. In the meanwhile, the autoencoder attempts to reconstruct the input image $x$ from the latent code vector $z$.

Different from GAN, in which the input noise lacks semantic information and the output distribution is uncontrollable, AAE largely increases the diversity of the output samples by making the latent code vector simulate the prior distribution. Therefore we can directly expand the training dataset through the generated samples by AAE without carefully selecting which image to use.

\subsubsection{Automatic labeling}
\begin{figure}[]
	\centering
	\includegraphics[width=3in]{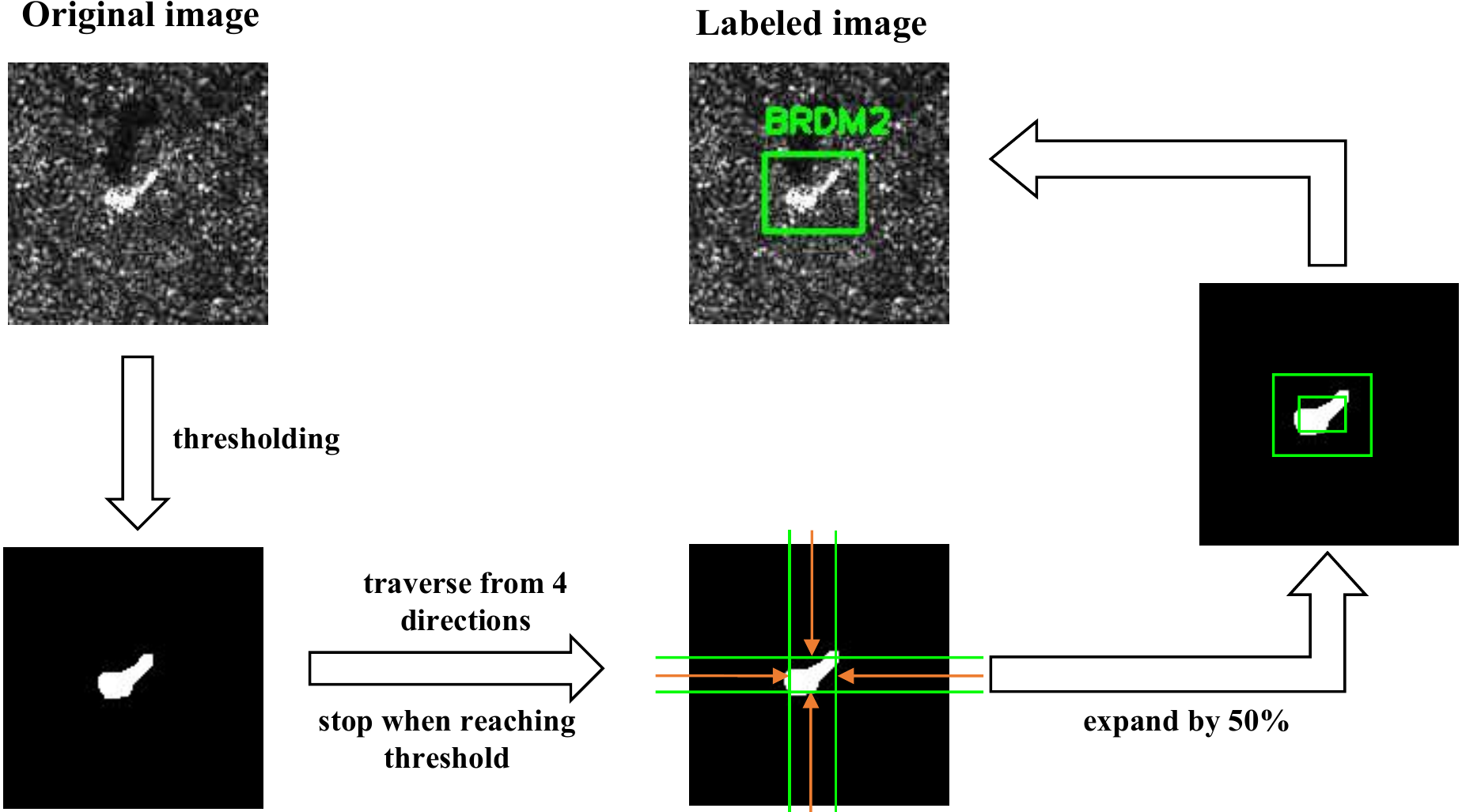}
	\caption{The method of automatic labeling.}
	\label{auto-label}
\end{figure}

After collecting all the training images, an automatic labeling method is developed to avoid manual labeling, which reduces a large amount of redundant work. The detailed design is shown in Algorithm \ref{auto_label}, and the process is shown in Fig. \ref{auto-label}.
\renewcommand{\algorithmicrequire}{ \textbf{Input:}} 
\renewcommand{\algorithmicensure}{ \textbf{Output:}} 

\begin{algorithm} 
	\caption{Automatic labeling.} 
	\label{auto_label} 
	\begin{algorithmic}[1] 
		\REQUIRE ~~\\ 
		The original SAR target chip.\\
		
		\ENSURE ~~\\
		The target coordinate $(x,y,w,h)$ and its corresponding category.
		
		\STATE Binarize the original image;\\
		
		\STATE Set a threshold of the white pixels' number and traverse the binarized image from 4 directions;\\
		\STATE Count the number of white pixels for each row or column;\\
		\STATE Stop traversing when reaching the threshold in each direction;\\
		\STATE Form the corresponding rectangle;\\
		\STATE Expand rectangle concentrically to a certain extent, e.g. 50\%, to produce a rectangle of proper size.
		
		\RETURN The target label information.
	\end{algorithmic}
\end{algorithm}

First, we need to binarize the original image using a thresholding method which will be introduced in Sec. \ref{2-2-1}. Though this binarization method can correctly segment the object, in a few cases there are still some small white spots in the background. To eliminate the effect of white spots in automatic labeling, we set a threshold of white pixel number to filter those spots and finally capture the object accurately. For each row or column, if the number of white pixels is lower than the threshold, we consider it not to constitute the target. We traverse the binarized image from 4 directions (up, down, left and right) to count the number of white pixels for each row or column, and stop traversing when reaching the threshold, formulating a rectangle which contains the center of the target. However, the edge of the target may also not reach the threshold thus may be filtered, so we need to expand the rectangle concentrically by a certain extent, which according to our experiment, is around 50\%, and the white pixel threshold of ten-class target usually lies in the interval [8,12].

\subsection{The ATR Framework on Complex Large-scene SAR Images}\label{2-2}
\begin{figure}[]
	\centering
	\includegraphics[width=3.5in]{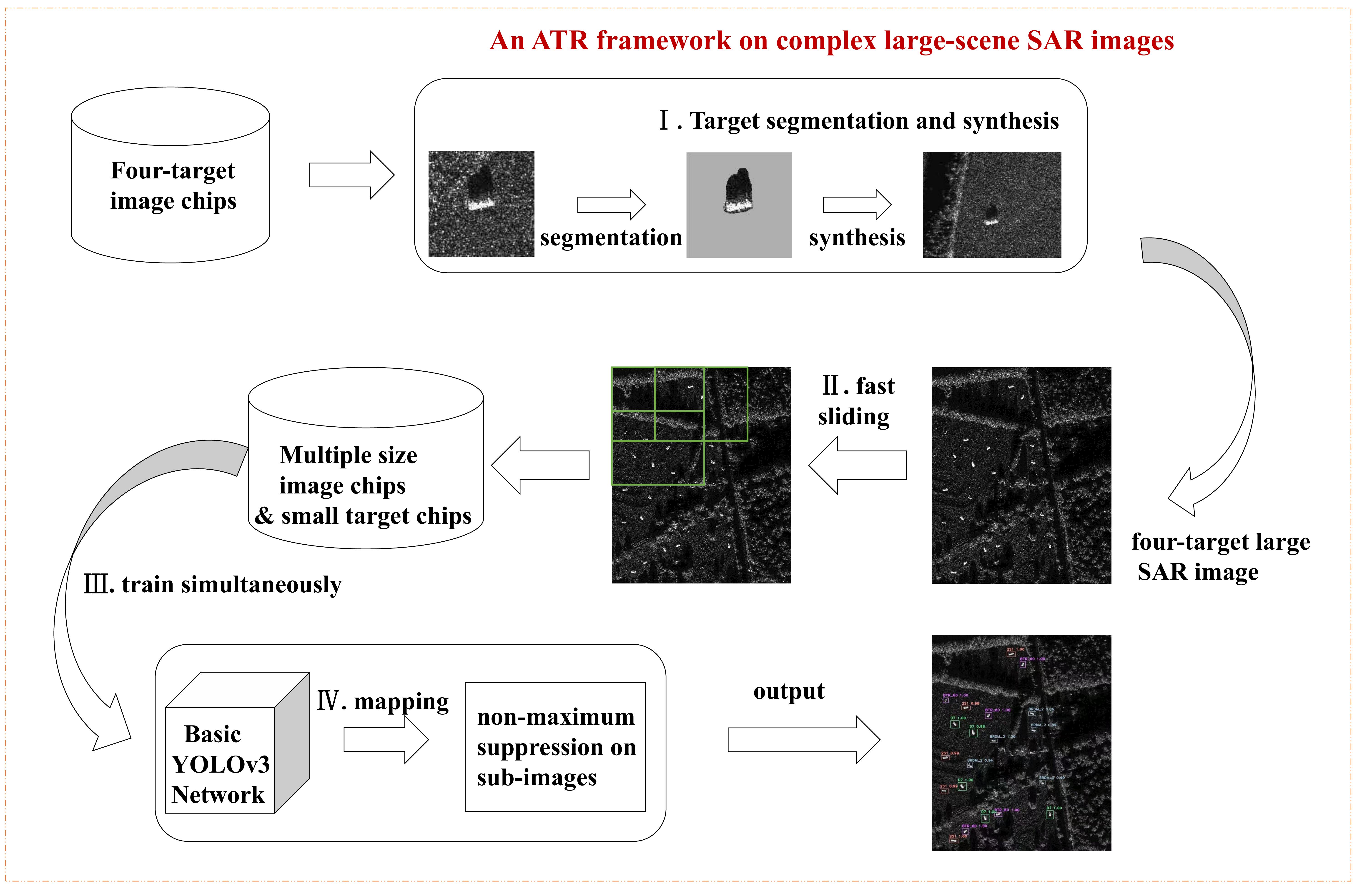}
	\caption{An ATR framework for complex large-scene SAR images. \uppercase\expandafter{\romannumeral1}. Target chips are segmented from their backgrounds and then synthesized on the large-scene SAR images. \uppercase\expandafter{\romannumeral2}. Fast sliding is conducted to divide the large-scene images into different sizes. \uppercase\expandafter{\romannumeral3}. Both sliced large-scene images and small target chips are fed into YOLOv3 network simultaneously to gain the final result. \uppercase\expandafter{\romannumeral3}. Finally we map the detection result to the large-scene image and apply non-maximum suppression to gain target with single bounding box.}
	\label{large-scene}
\end{figure}

After obtaining labeled small target chips with data augmentation, in this part, we will apply our ATR framework to detect multiple targets on complex large-scene SAR images. As is shown in Fig. \ref{large-scene}, firstly to prepare the large-scene database, we need to segment the target and its shadow from the speckle background; then the target is synthesized into the large-scene background which is acquired under the same depression degree; later a fast sliding method is used to divide the synthesized large-scene image into different sizes and a YOLOv3 network is adopted to train large-scene images and small target chips simultaneously to gain the final result on large-scene images with target categories, confidences and bounding boxes.

\begin{figure}[]
	\centering
	\includegraphics[width=2.5in]{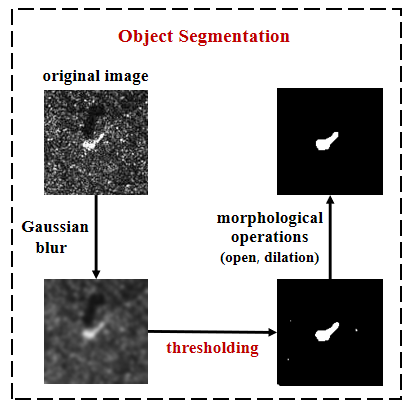}
	\caption{Object segmentation process. We sequentially adopt Gaussian blur, thresholding and morphological operation to generate the final segmented object. The threshold value is set as: $p=121$.}
	\label{object_segmentation}
\end{figure}
\subsubsection{Database preparation}\label{2-2-1}
Since it is costly to directly obtain large-scene SAR images with multiple targets, a target segmentation and synthesis method is first proposed to establish a training database.

\begin{figure}[]
	\centering
	\includegraphics[width=3in]{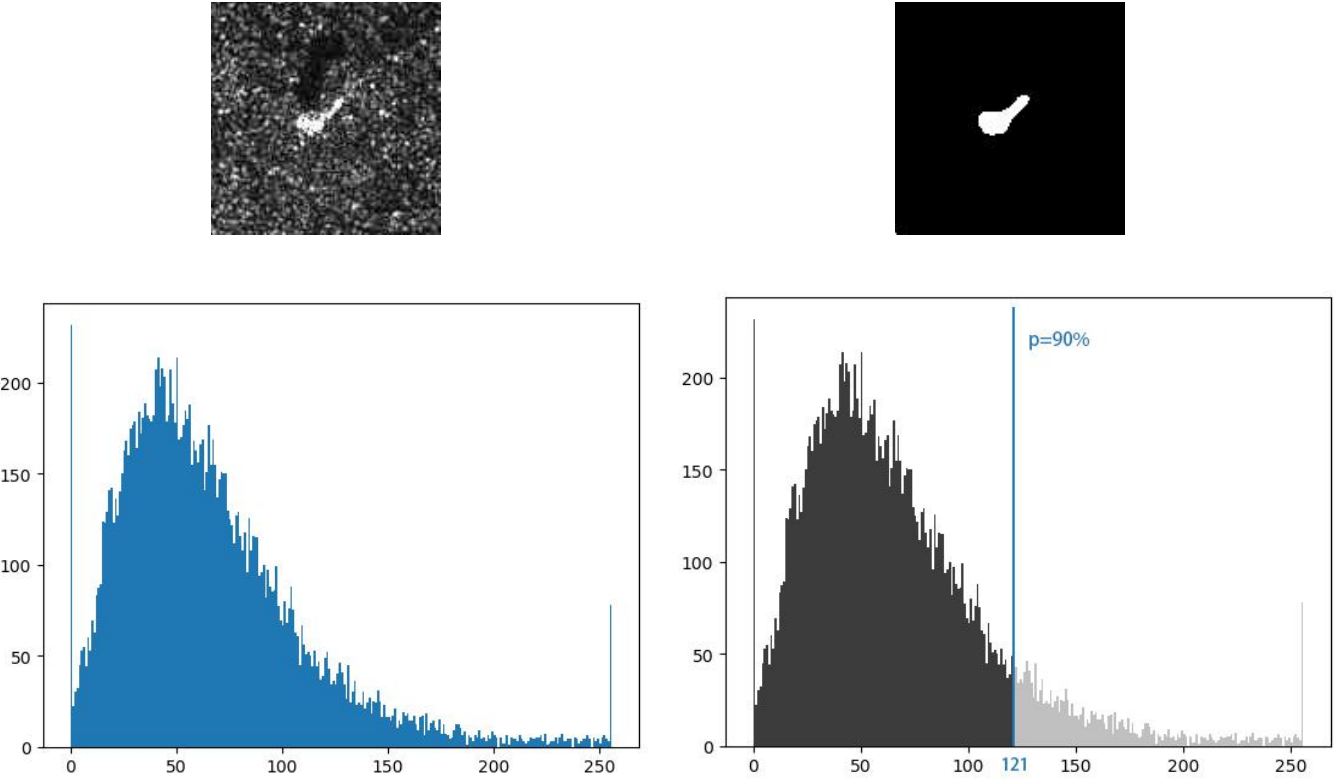}
	\caption{The pixel value distribution of the SAR image. The left column is the original image and its histogram, and the right column is the result of applying our selected threshold $p=121$.}
	\label{threshhold}
\end{figure}

For the target segmentation part, we segment the object and its shadow in two steps, which are represented in Fig. \ref{object_segmentation}. The first step is to segment the object without its shadow. The image is firstly smoothed by Gaussian blur, with the convolution kernel determined adaptively by the image itself. A blurred image has fewer noises and then can be binarized. The selection of the threshold is semi-adaptive and it is the most critical step. Since there is only one target lying in the approximate center in the SAR target chip and it is obviously distinguished from the background, the binarization rule is set as:
\begin{equation}
(x',y') = \left\{ \begin{array}{l}
255\quad if\, (x,y) > p\\
0\quad\quad otherwise
\end{array} \right.
\end{equation} 
where $p$ denotes the threshold.                      
The threshold can be determined by the pixel value proportion of the target in the whole image. The value $p$ is selected around 90\% in most circumstances, which can be estimated from the intuitive area ratio. As is shown in Fig. \ref{threshhold}, the object takes up about 10\% of the image, so we choose $p=121$ in this example as the threshold and apply the binarization rule. This method proves its effectiveness on MSTAR target chips since it can successfully separate more than 95\% objects in the original images.

\begin{figure}[]
	\centering
	\includegraphics[width=3in]{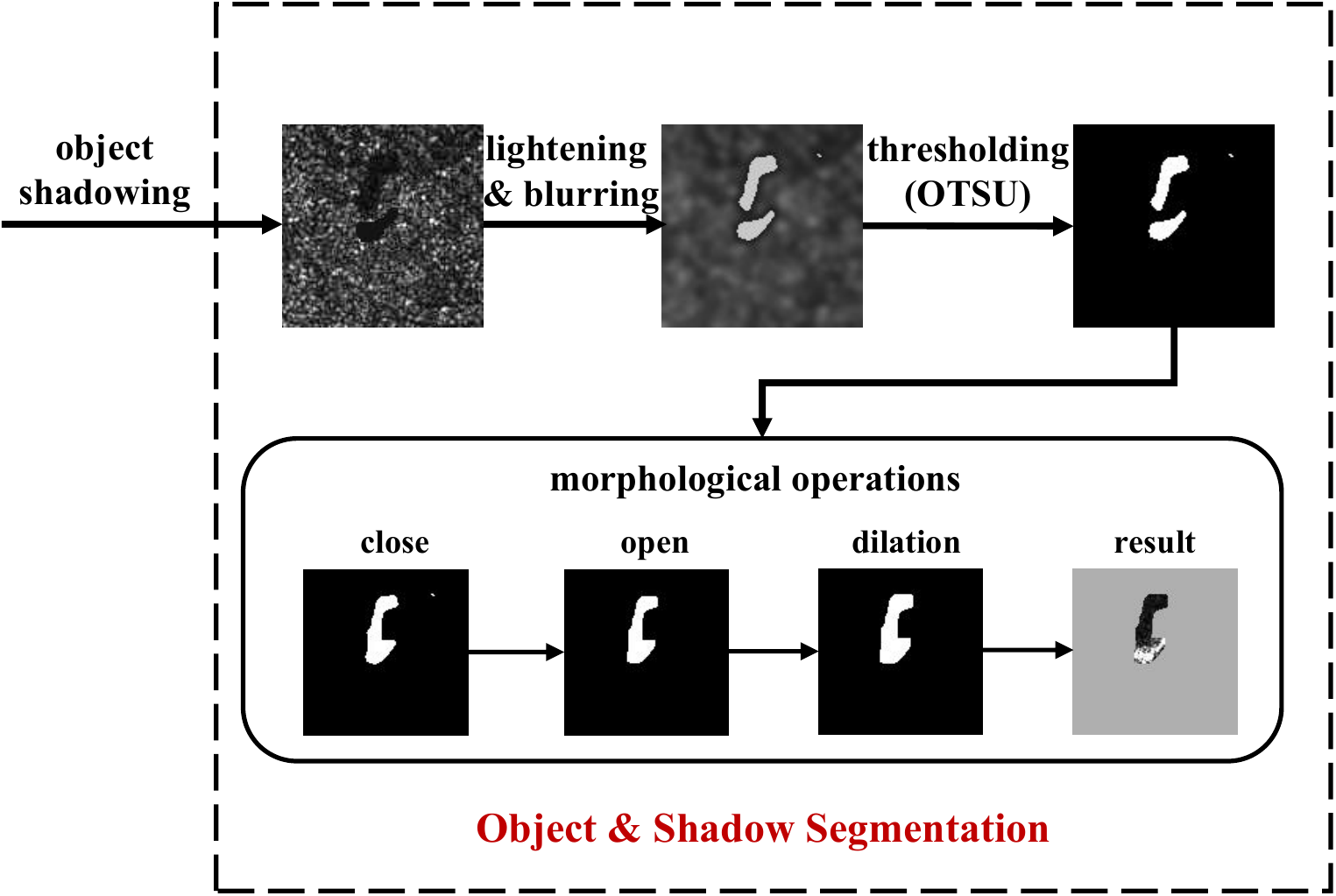}
	\caption{Object and shadow segmentation process. The object segmented from step one will be processed through four procedures: object shadowing, lightening and blurring, thresholding, and final morphological operations.}
	\label{ob_shsyn}
\end{figure}

In the second step, the object and its shadow are segmented in the meantime, which is exhibited in Fig. \ref{ob_shsyn}. Firstly the segmented object from step one is shadowed to a comparatively small pixel value, then the binarization rule is adopted to highlight both the object and its shadow. After Gaussian blur, an adaptive threshold selection algorithm OTSU is adopted to segment them directly, since the present image has low noise. Then morphological operations are used to improve the segmentation result.

Usually the object and its shadow are separated after binarization, so the closing operation is first adopted for connection, and then the opening operation is applied to clear up small spots while keeping the main body unchanged, at last we add some edge details by dilation to produce the final segmentation result. It is clear to see that the object and its shadow are successfully segmented using our method.

\begin{figure}[]
	\centering
	\includegraphics[width=3in]{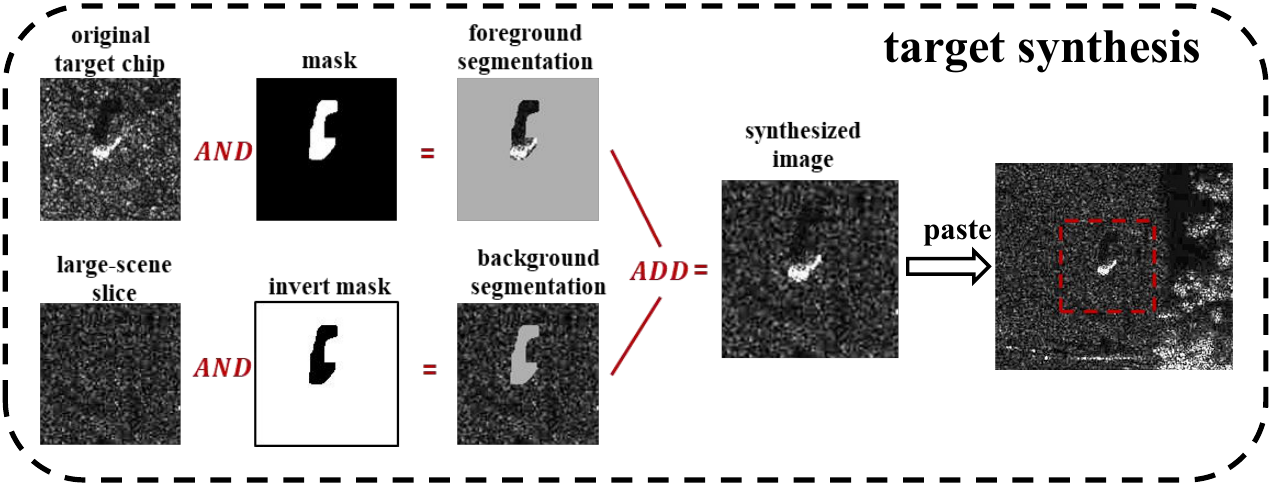}
	\caption{Target synthesis process. Note that the original target chip and the cutted large-scene slice are of the same size.}
	\label{ob_sh}
\end{figure}

After segmentation, it is of the same importance to synthesize the large-scene image with multiple targets naturally. As is shown in Fig. \ref{ob_sh}, the target synthesis process also goes through two steps. The first one is to design and record the target distribution in the large-scene SAR image. We randomly select 20 coordinates to put four-class targets and each class targets occupy five positions. Since there are some obstacles such as trees, buildings and rivers in the background, and the shadows of these obstacles have a certain direction due to the specific shooting angle and time, it is necessary to carefully select those targets which have the similar direction of shadow and finally design the location of the targets to ensure that they will not fall into the obstacles. 
After choosing the proper target and its coordinate, the second step is to cut a slice of the large-scene image in the corresponding designed position, which has the same shape as the target chip, then the mask produced by target segmentation and its inverse mask are used to perform the bitwise-and operation on the original image and the scene cut respectively. Finally these two operation results are combined to get the final natural synthesized large-scene image.
\subsubsection{Fast sliding}
\begin{figure}[]
	\centering
	\includegraphics[width=3.5in]{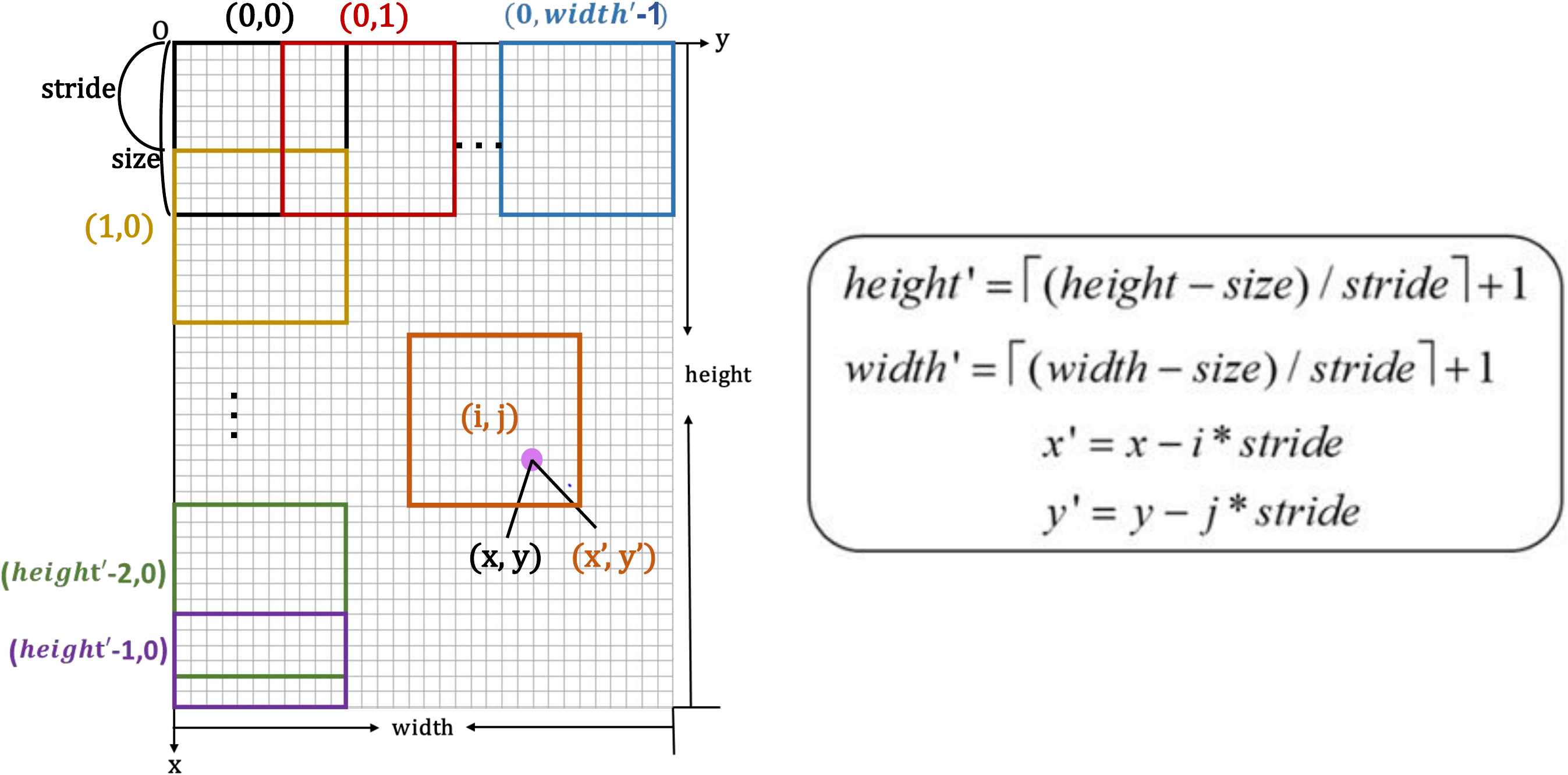}
	\caption{The proposed fast sliding method. 
		$(x,y)$ represents the target position in the large-scene image; $(x',y')$ denotes the corresponding coordinate in the cropped slice.}
	\label{fast_sliding}
\end{figure}

Since the number of the large-scene images in the MSTAR dataset is limited, directly feed them into the YOLOv3 network will cause severe overfitting problems. Therefore, we can use a fast sliding method to expand the training dataset while still containing complex background information. However, if we randomly set the size of the sliding window and its stride, the target in the scene is likely to be divided into several parts and we may also obtain a large number of slices without targets which will increase the input data redundancy. Therefore, a fast sliding method is proposed which uses sliding windows to cut a synthesized large-scene image into small slices to expand the input data volume. Different from the method in work\cite{cui2019d}, we do not need to make sure that the size of the sliding window is fixed, so when the sliding window almost reaches the edge of the large-scene image, the remaining part which is smaller than the setted size can be directly cut from the image and saved for training. The fast sliding method is shown in Fig. \ref{fast_sliding}.

When calculating the corresponding target coordinate in the cropped slice, we respectively use
$height'$ and $width'$ to represent the number of slices that can be obtained from the vertical and horizontal direction (including the last possible incomplete slice). $(i,j)$ denotes the coordinate of a slice in the large-scene image, while $(x,y)$ along with $(x',y')$ respectively denote the coordinate of a target in the large-scene image and the cropped slice. For each sliding window $(i,j)$, we traverse the coordinates of the targets and if there is any target which could meet the following conditions simultaneously, we consider it falling into this sliding window completely.
\begin{equation}
\begin{aligned}
&{x_{\min }} > i * stride\\
&{x_{\max }} < i * stride + size\\
&{y_{\min }} > j * stride\\
&{y_{\max }} < j * stride + size
\end{aligned}
\end{equation}
Meanwhile, the slice will be automatically abandoned if there is no target falling into it or the target is incomplete and the sliding window will move on to the next one until the whole image is covered. 
In this paper, we choose four sizes $128 \times 128$, $256 \times 256$, $512 \times 512$, $1024 \times 1024$ to apply fast sliding.

\subsubsection{Training on YOLOv3 network}
With expanded small target chips and large-scene image slices, we feed them into the YOLOv3 network using pre-trained weight on the COCO dataset. The main idea of YOLO is to divide the input image into
$S \times S$ grids and if the center of an object falls into one grid cell, then that grid cell is responsible for predicting the object.

\begin{figure}[]
	\centering
	\includegraphics[width=3.5in]{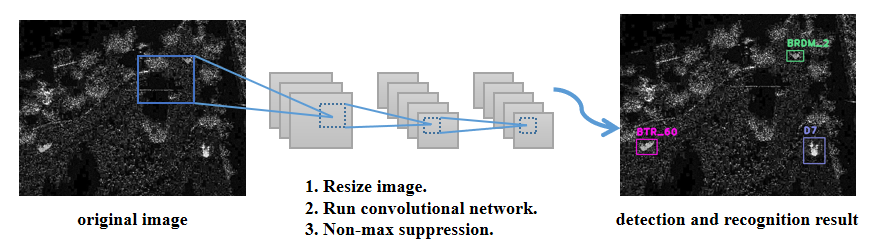}
	\caption{The basic YOLO detection system.}
	\label{yolo}
\end{figure}

Confidence is defined as
$\Pr (Object) * IOU_{pred}^{truth}$, which reflects how confident the model is that the bounding box contains an object and indicates the accuracy of the prediction.
$\Pr (Clas{s_i}\left| {Object)} \right.$  denotes the probabilities that each grid cell predicts $C$ classes. 
\begin{equation}
\label{eqn_example2}
\begin{aligned}
\Pr (Clas{s_i}|Object) * \Pr (Object) * IOU_{pred}^{truth} \\
= \Pr (Clas{s_i}) * IOU_{pred}^{truth}
\end{aligned}
\end{equation}

By multiplying these two parts, we can obtain both the probability of that class appearing in the box and how well the predicted box fits the object. The loss function is defined as:
{\small
\begin{equation}
\centering
\begin{aligned}
loss = &{\lambda _{coord}}\sum\limits_{i = 0}^{{S^2}} {\sum\limits_{j = 0}^B { ]\kern-0.15em] _{ij}^{obj}[({x_i} - {{\hat x}_i}} } {)^2} + {({y_i} - {\hat y_i})^2}]\\
+ &{\lambda _{coord}}\sum\limits_{i = 0}^{{S^2}} {\sum\limits_{j = 0}^B { ]\kern-0.15em] _{ij}^{obj}[(\sqrt {{\omega _i}}  - \sqrt {{{\hat \omega }_i}} } } {)^2} + {(\sqrt {{h_i}}  - \sqrt {{{\hat h}_i}} )^2}]\\
+ &\sum\limits_{i = 0}^{{S^2}} {\sum\limits_{j = 0}^B { ]\kern-0.15em] _{ij}^{obj}{{({C_i} - {{\hat C}_i})}^2}} }+ \sum\limits_{i = 0}^{{S^2}} { ]\kern-0.15em] _i^{obj}\sum\limits_{c \in classes} {{{({p_i}(c) - {{\hat p}_i}(c))}^2}} } \\
 + &{\lambda _{noobj}}\sum\limits_{i = 0}^{{S^2}} {\sum\limits_{j = 0}^B { ]\kern-0.15em] _{ij}^{noobj}{{({C_i} - {{\hat C}_i})}^2}} }   
\end{aligned}
\end{equation}
}
where ${ ]\kern-0.15em] _i^{obj}}$ denotes whether the object appears in cell $i$, and 
${ ]\kern-0.15em] _{ij}^{obj}}$ denotes that the $jth$ bounding box in cell $i$ is responsible for the prediction. $\lambda _{coord}$ is used to increase the loss from bounding box coordinate predictions and $\lambda _{noobj}$ is to decrease the loss from confident predictions for boxes that don't contain objects. YOLOv3 has represented superior performance on both accuracy and speed. Compared with Faster R-CNN, which needs to repeatedly train the region proposal network (RPN) and Fast R-CNN, YOLOv3 is much faster without training on RPN and just need to "Look Once" to obtain both the location and classification of the object. As to SSD, which is also fast but inferior on small target detection due to low semantic value for the bottom layer, YOLOv3 is even faster based on the logistic loss while containing the competitive accuracy of detecting small targets since higher resolution layers can obtain higher semantic values. 

When the training process is finished, each detected object's bounding box information, including (coordinates$(x,y,w,h)$, $class$ and $confidence$), is automatically set down for non-maximum suppression.

\begin{figure}[]
	\centering
	\includegraphics[width=3in]{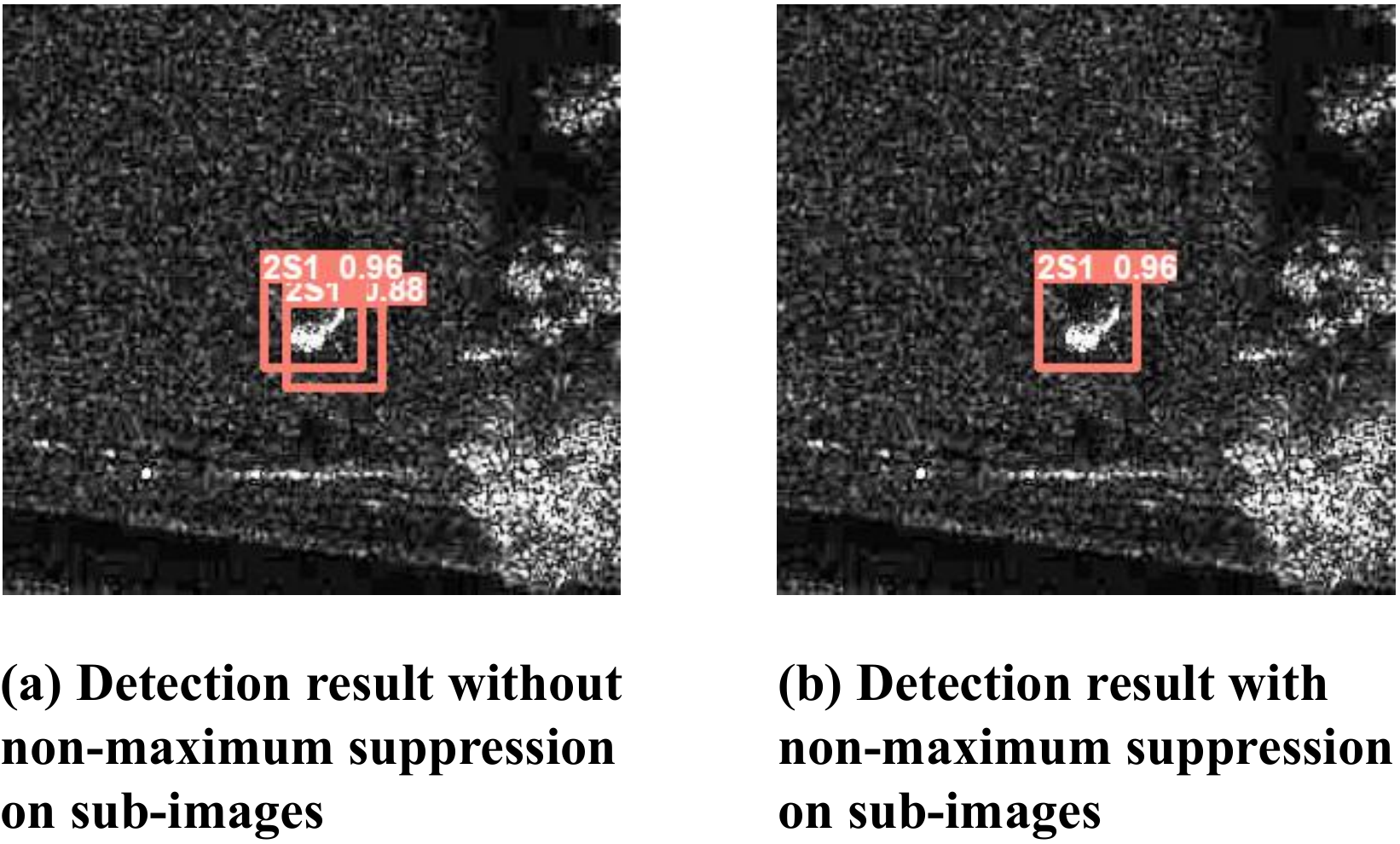}
	\caption{Non-maximum suppression on sub-images for one target. The left column is  the detection result without non-maximum suppression on sub-images; the right column is the detection result with non-maximum suppression on sub-images.}
	\label{NMSS}
\end{figure}

To conduct non-maximum suppression on sub-images, firstly, the bounding boxes information of targets on each sub-image is mapped to the large-scene image using coordinate conversion, and then non-maximum suppression is applied to the targets with multiple bounding boxes, which means the bounding box with the highest score remains and other boxes which have $IOU > 0.7$ with it are deleted. Fig. \ref{NMSS} shows how it works on one target. In the end, we can easily obtain large-scene images with multiple targets detected and recognized. 

\section{Experimental Results}

\label{3}
\subsection{MSTAR Dataset}
\begin{figure}[]
	\centering
	\includegraphics[width=3.5in]{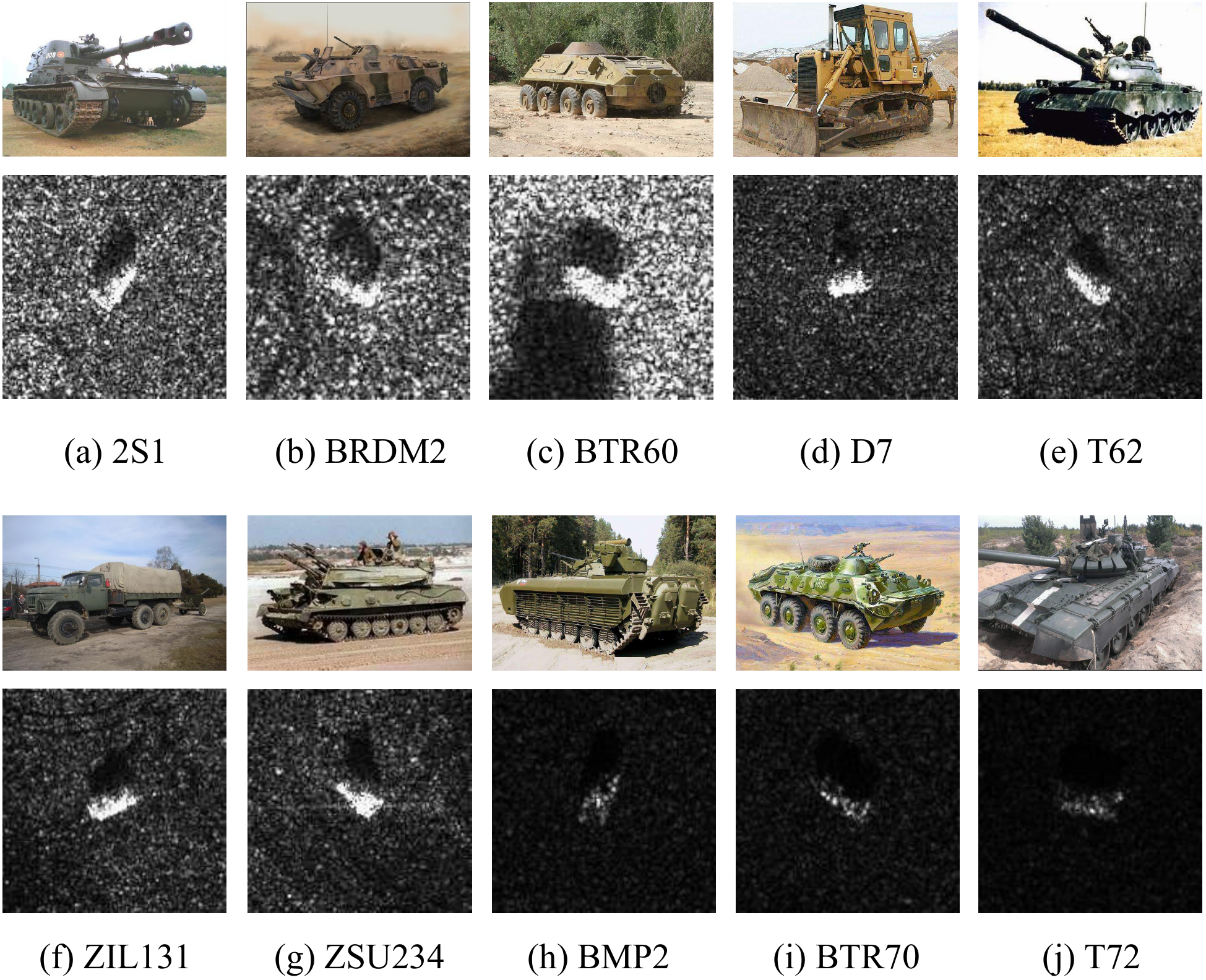}
	\caption{Types of military targets: (top) optical images versus (bottom) SAR images.}
	\label{MSTAR}
\end{figure}

We use the MSTAR dataset to complete our experiments. The MSTAR dataset includes thousands of SAR images, including ten categories of ground military vehicles (armored personnel carrier: BMP2, BRDM2, BTR60, and BTR70; tank: T62 and T72; rocket launcher: 2S1; air defense unit: ZSU234; truck: ZIL131; and bulldozer: D7). They were collected under an X-band SAR sensor, in a 1-ft resolution spotlight mode, full aspect coverage (in the range of $0^{\circ}$ to $360^{\circ}$). The MSTAR dataset is widely used to test the performance of a SAR-ATR system. Fig. \ref{MSTAR} shows the optical images and the corresponding SAR images. The number of images for training in our experiment is summarized in Table \ref{mstar}. Besides small target chips, the MSTAR dataset also provides simple and complex scene images without targets, and these backgrounds include river, sea surface, forest and so on. 

\begin{table}[]\footnotesize
	\centering
	\caption{MSTAR training and testing dataset.}
	\label{mstar}
	\begin{tabular}{ccccc}
		\toprule[2pt]
		\multirow{2}{*}{Targets} & \multicolumn{2}{c}{Train} & \multicolumn{2}{c}{Test} \\ \cline{2-5} 
		& No.Images   & Depression  & No.Images  & Depression  \\ \hline
		2S1                      & 299         & 17${\rm{^\circ }}$         & 274        & 15${\rm{^\circ }}$         \\
		BRDM2                    & 298         & 17${\rm{^\circ }}$         & 274        & 15${\rm{^\circ }}$         \\
		BTR60                    & 256         & 17${\rm{^\circ }}$         & 195        & 15${\rm{^\circ }}$         \\
		D7                       & 299         & 17${\rm{^\circ }}$         & 274        & 15${\rm{^\circ }}$         \\
		T62                      & 299         & 17${\rm{^\circ }}$         & 273        & 15${\rm{^\circ }}$         \\
		ZIL131                   & 299         & 17${\rm{^\circ }}$         & 274        & 15${\rm{^\circ }}$         \\
		ZSU234                   & 299         & 17${\rm{^\circ }}$         & 274        & 15${\rm{^\circ }}$         \\
		BMP2                     & 233         & 17${\rm{^\circ }}$         & 196        & 15${\rm{^\circ }}$         \\
		BTR70                    & 233         & 17${\rm{^\circ }}$         & 196        & 15${\rm{^\circ }}$         \\
		T72                      & 232         & 17${\rm{^\circ }}$         & 196        & 15${\rm{^\circ }}$   \\
		\bottomrule[2pt]     
	\end{tabular}
\end{table}
\subsection{Experiment of Automatic Labeling}

We use the MSTAR ten-class training dataset to conduct the experiment of automatic labeling. The result of the automatic labeling method is shown in Table \ref{label-mstar}. We can see that the average error rate is 1.15\%, and six-class targets including D7, T62, ZSU234, BMP2, BTR70 and T72 are perfectly labeled without missing or not correctly marked, which proves the effectiveness and efficiency of our approach.
\begin{table}[]\footnotesize
\centering
\caption{automatic labeling result on MSTAR ten-class targets.}
\label{label-mstar}
\begin{tabular}{cccc}
\toprule[2pt]
\textbf{Class}   & \textbf{Image Num} & \textbf{Not correctly labeled} & \textbf{Error rate/(\%)} \\ \hline
\textbf{2S1}     & 299                & 10                             & 3.34                     \\
\textbf{BRDM2}   & 298                & 6                              & 2.01                     \\
\textbf{BTR60}   & 256                & 3                              & 1.17                     \\
\textbf{D7}      & 299                & 0                              & 0                        \\
\textbf{T62}     & 299                & 0                              & 0                        \\
\textbf{ZIL131}  & 299                & 15                             & 5.02                     \\
\textbf{ZSU234}  & 299                & 0                              & 0                        \\
\textbf{BMP2}    & 233                & 0                              & 0                        \\
\textbf{BTR70}   & 233                & 0                              & 0                        \\
\textbf{T72}     & 232                & 0                              & 0                        \\
\textbf{Average} & -                  & -                              & \textbf{1.15}            \\ 
\bottomrule[2pt]
\end{tabular}
\end{table}
\subsection{Detection and Recognition on SAR Target Chips}
In this part, we firstly conduct some experiments on the small SAR target chips to show the effectiveness of the AAE data augmentation method. Since the targets locate right in the center of the image chips, the result can only consist of three aspects: target not detected, target not correctly detected and target correctly detected. Therefore we will use accuracy (ACC) and False Negative Rate (FNR) as indicators, which shows how many targets we have missed and not correctly detected during detection. ACC and FNR are respectively defined as:
\begin{equation}
\label{ACC}
ACC = \frac{{TP}}{{TP + FN}}
\end{equation}
\begin{equation}
\label{FNR}
FNR = \frac{{FN}}{{TP + FN}}
\end{equation}
FN denotes the number of not correctly detected targets and the missing targets; TP denotes the number of correctly detected targets.
\subsubsection{Experiment without data augmentation}
The first experiment is conducted under the YOLOv3 framework without data augmentation, aiming to detect and classify these ten targets. The total 2747 image chips acquired under 17${\rm{^\circ }}$ depression angle are used for training and the 2426 image chips obtained under 15${\rm{^\circ }}$ are tested. 

Our YOLOv3 network parameters are shown as follow: $anchors = 10$, $14$, $23$, $27$, $37$, $58$, $81$, $82$, $135$, $169$, $344$, $319$; $class = 10$; $ignore\_thresh = 0.7$;
$true\_thresh = 1$; $random = 1$, and the basic network parameters are set as: $batch\_size = 64$; $subdivisions = 2$; $momentum = 0.9$; $decay = 0.0005$; $learning\_rate = 0.001$; $epoch =200$. Besides that, we use the pretrained weights on COCO image set, and then feed SAR images into our network.

As is presented in Table \ref{without_aae}, the worst accuracy is 92.31\%, which is caused by T62 since 19 targets are recognized as ZSU234. Besides, targets 2S1, BTR60 and BMP2 are also not well correctly detected and classified. 
\begin{table*}[]
	\centering
	\caption{Confusion matrix for ten-class SAR image detection and recognition without AAE.}
	\label{without_aae}
	\resizebox{\textwidth}{!}
	{
		\begin{tabular}{cccccccccccccc}
			\toprule[2pt]
			\textbf{class}   & \textbf{2S1} & \textbf{BRDM2} & \textbf{BTR60} & \textbf{D7} & \textbf{T62} & \textbf{ZIL131} & \textbf{ZSU234} & \textbf{BMP2} & \textbf{BTR70} & \textbf{T72} & \textbf{None} & \textbf{ACC(\%)} & \textbf{FNR(\%)} \\ \hline
			\textbf{2S1}     & 269          & 0              & 0              & 0           & 2            & 3               & 0               & 0             & 0              & 0            & 0             & 98.18            & 1.82                \\
			\textbf{BRDM2}   & 0            & 271            & 0              & 1           & 0            & 1               & 1               & 0             & 0              & 0            & 0             & 98.91            & 1.09                \\
			\textbf{BTR60}   & 0            & 1              & 186            & 0           & 0            & 0               & 4               & 0             & 0              & 0            & 4             & 95.38            & 4.62             \\
			\textbf{D7}      & 0            & 0              & 0              & 268         & 0            & 0               & 6               & 0             & 0              & 0            & 0             & 97.81            & 2.19                \\
			\textbf{T62}     & 1            & 0              & 0              & 0           & 252          & 0               & 19              & 0             & 0              & 0            & 1             & 92.31            & 7.98             \\
			\textbf{ZIL131}  & 0            & 0              & 0              & 0           & 0            & 274             & 0               & 0             & 0              & 0            & 0             & 100              & 0                \\
			\textbf{ZSU234}  & 0            & 0              & 0              & 0           & 0            & 0               & 274             & 0             & 0              & 0            & 0             & 100              & 0                \\
			\textbf{BMP2}    & 0            & 0              & 0              & 0           & 0            & 0               & 0               & 191           & 0              & 5            & 0             & 97.45            & 2.55                \\
			\textbf{BTR70}   & 0            & 0              & 0              & 0           & 0            & 0               & 0               & 0             & 196              & 0            & 0             & 100              & 0                \\
			\textbf{T72}     & 0            & 0              & 0              & 0           & 0            & 0               & 0               & 0           & 0              & 196            & 0             & 100              & 0                \\
			\textbf{Average} & \multicolumn{11}{c}{}                                                                                                                                                           & \textbf{97.98}            & \textbf{2.025}             \\ \bottomrule[2pt]
		\end{tabular}
	}
\end{table*}
\subsubsection{Experiment on different generative networks}
\begin{figure}[]
	\centering
	\includegraphics[width=3.2in]{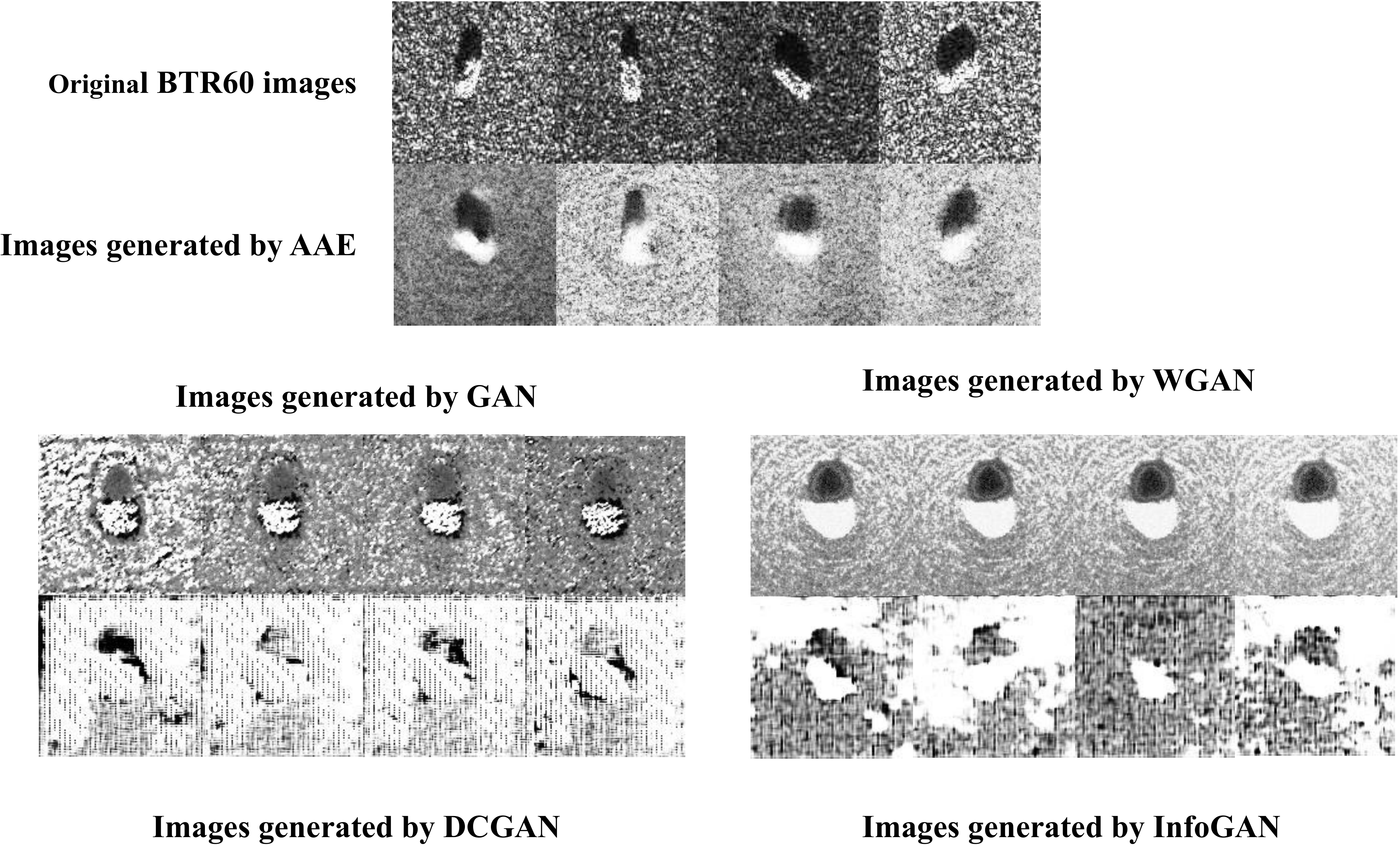}
	\caption{Generated SAR images based on different generative models.}
	\label{aae_gan}
\end{figure}

In order to improve the classification accuracy, we adopt AAE to expand our dataset. We choose BTR60 (size $128 \times 128$) and set 200 training epochs in this experiment to illustrate the effectiveness of the AAE method. Fig. \ref{aae_gan} shows the generated SAR images under different generative models. To further illustrate the effectiveness of the AAE model, we use Fréchet Inception Distance (FID) \cite{Dowson1982The} to further evaluate the variety of the generated objection, which is defined in Eq. (\ref{gan}). The FID score of different generative models is shown in Table \ref{4}.
\begin{equation}\label{gan}
FID{\rm{ = ||}}{\mu _{\rm{r}}}{\rm{ - }}{\mu _{\rm{g}}}{\rm{|}}{{\rm{|}}^2}{\rm{ + }}Tr(\sum\nolimits_r { + \sum\nolimits_g { - 2(\sum\nolimits_r {\sum\nolimits_g {{)^{1/2}})} } } }
\end{equation}
where ${\mu _{\rm{r}}}$, ${\mu _{\rm{g}}}$ and $\sum\nolimits_r $, $\sum\nolimits_g $ are the respective means and covariance matrices of real and generated images.

\begin{table}[]\footnotesize
	\centering
	\caption{FID score on different generative models.}
	\begin{tabular}{llllll}
		\toprule[2pt]
		\textbf{}    & \textbf{AAE}     & \textbf{GAN}     & \textbf{DCGAN}   & \textbf{WGAN}    & \textbf{InfoGAN} \\ \hline
		\textbf{FID} & 195.943 & 226.687 & 401.827 & 353.254 & 399.678 \\ 
		\bottomrule[2pt]
		\label{fid}
	\end{tabular}
\end{table}

\begin{table*}[]
	\centering
	\caption{Confusion matrix for ten-class SAR image detection and recognition with AAE.}
	\resizebox{\textwidth}{!}
	{
		\begin{tabular}{@{}cccccccccccccc@{}}
			\toprule[2pt]
			\textbf{class}   & \textbf{2S1} & \textbf{BRDM2} & \textbf{BTR60} & \textbf{D7} & \textbf{T62} & \textbf{ZIL131} & \textbf{ZSU234} & \textbf{BMP2} & \textbf{BTR70} & \textbf{T72} & \textbf{None} & \textbf{ACC(\%)} & \textbf{FNR(\%)} \\ \midrule
			\textbf{2S1}     & 273          & 0              & 0              & 0           & 0            & 1               & 0               & 0             & 0              & 0            & 0             & 99.64            & 0.36                \\
			\textbf{BRDM2}   & 0            & 272            & 0              & 0           & 0            & 1               & 1               & 0             & 0              & 0            & 0             & 99.27            & 0.72                \\
			\textbf{BTR60}   & 1            & 2              & 184            & 0           & 0            & 0               & 5               & 0             & 0              & 0            & 3             & 95.83            & 5.64             \\
			\textbf{D7}      & 0            & 0              & 0              & 271         & 0            & 0               & 3               & 0             & 0              & 0            & 0             & 98.91            & 1.09                \\
			\textbf{T62}     & 2            & 0              & 0              & 0           & 268          & 1               & 1               & 0             & 0              & 0            & 1             & 98.17            & 1.83             \\
			\textbf{ZIL131}  & 0            & 0              & 0              & 1           & 0            & 273             & 0               & 0             & 0              & 0            & 0             & 99.64            & 0.36                \\
			\textbf{ZSU234}  & 0            & 0              & 0              & 0           & 0            & 0               & 274             & 0             & 0              & 0            & 0             & 100              & 0                \\
			\textbf{BMP2}    & 0            & 0              & 0              & 0           & 0            & 0               & 1               & 193           & 0              & 2            & 0             & 98.47            & 1.53                \\
			\textbf{BTR70}   & 0            & 0              & 0              & 0           & 0            & 0               & 0               & 0             & 196              & 0            & 0             & 100              & 0                \\
			\textbf{T72}     & 0            & 0              & 0              & 0           & 0            & 0               & 0               & 1             & 0              & 195            & 0             & 99.49            & 0.51                \\
			\textbf{Average} & \multicolumn{11}{c}{}                                                                                                                                                           & \textbf{98.89}            & \textbf{1.204}             \\ \bottomrule[2pt]
			\label{with_aae}
		\end{tabular}
	}
\end{table*}
\subsubsection{Experiment with data augmentation}
To evaluate the effectiveness of this data augmentation method, we implement it in the ten-class SAR target task. 
The number of images generated for training will strongly influence the classification results and a better classification result emerges when the number of generated images equals to half of the number of original images \cite{he2019parallel}. However, it is not sufficient enough to derive the most proper ratio in order to gain the best classification result.
In our experiment, the targets, which are more likely to be recognized by mistake, are 2S1, BTR60, D7 and T62. So we first separately generated 100 images to add to the four-target dataset. However, BTR60 and T62 still occupy a high error rate, thus additional 100 generated images are added into BTR60 and T62 dataset. The final result shows that the accuracy of BTR60 increases by 0.45\%, while the accuracy of T62 raises up by 5.86\%. When we continuously increase the generated image proportion, the outcome does not improve. 

As is shown in Table \ref{with_aae},  the average accuracy rate reaches 98.89\%, while the FNR is controlled around 1.2\%. The main error is caused by ZSU234,  since some images of BTR60 and D7 are recognized as ZSU234 by mistake. 

\subsubsection{Comparison among different ATR methods}
\begin{figure}[]
	\centering
	\includegraphics[width=3.5in]{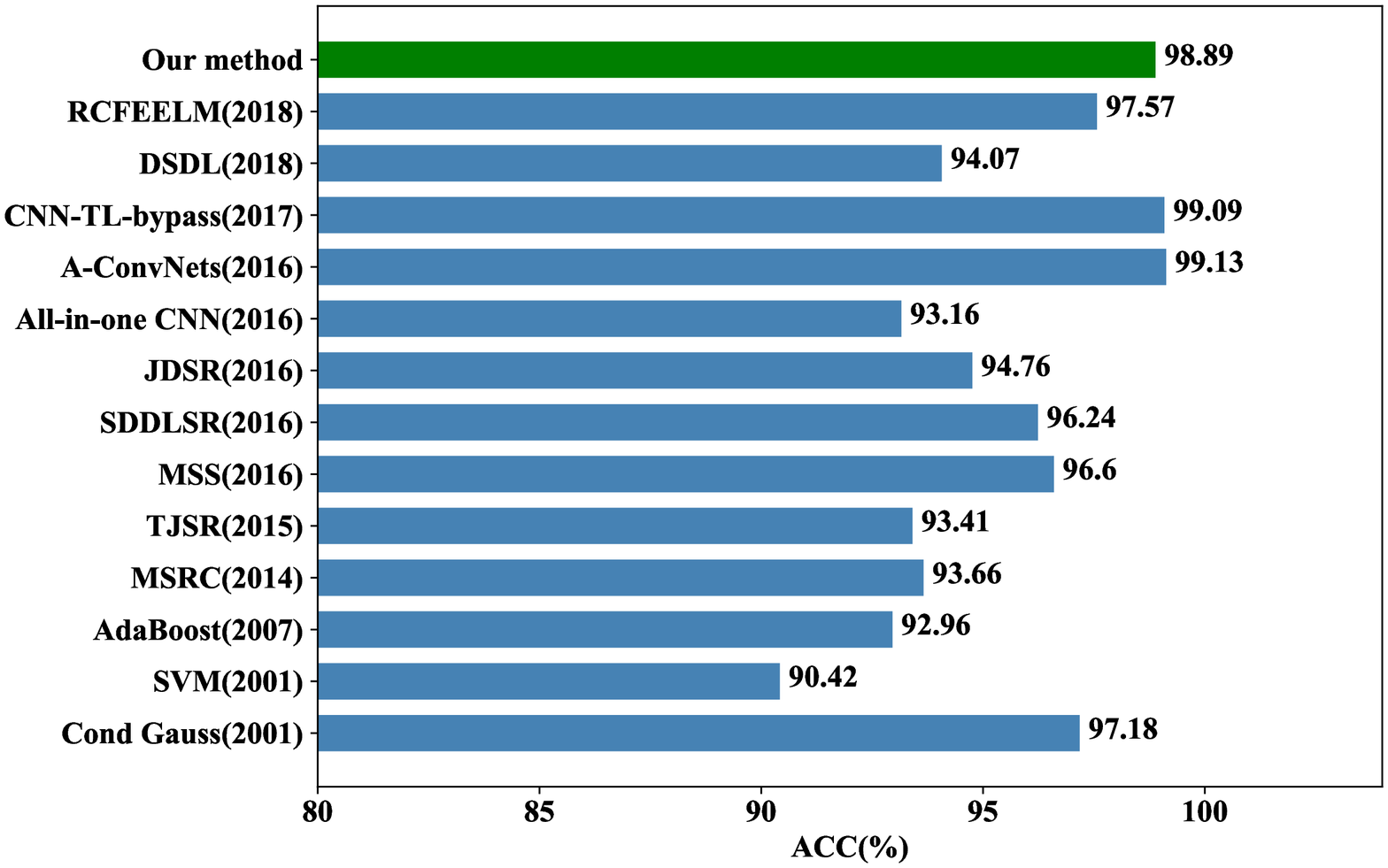}
	\caption{The classification accuracy of proposed method versus some previous methods and state-of-the-art methods.}
	\label{compare}
\end{figure}

Fig. \ref{compare} illustrates the performance of the proposed method compared to other SAR recognition methods. It has been shown that our method outperforms many methods, i.e., conditionally gaussian model (Cond Gauss) \cite{O2001SAR}, support vector machines (SVM) \cite{Zhao2001Support}, adaptive boosting (AdaBoost) \cite{Sun2007Adaptive}, sparse representation of monogenic signal (MSRC) \cite{NoneSparse}, monogenic scale space (MSS) \cite{DongClassification}, tri-task joint sparse representation (TJSR) \cite{Dong2017SAR}, supervised discriminative dictionary learning and sparse representation (SDDLSR) \cite{song2016sar}, joint dynamic sparse representation (JDSR) \cite{Sun2016SAR} and All-in-one CNN \cite{ding2016convolutional}. In addition, our method achieves a competitive performance compared to the state-of-the-art methods A-ConvNets \cite{chen2016target}, CNN-TL-bypass \cite{huang2017transfer}, discriminative statistical dictionary learning (DSDL) \cite{liu2018sar}, random convolution features and ensemble extreme learning machines (RCFEELM) \cite{Gu2018Fast}.
\subsubsection{Experiment of anti-noise performance}

\begin{figure}[]
	\centering
	\includegraphics[width=3.5in]{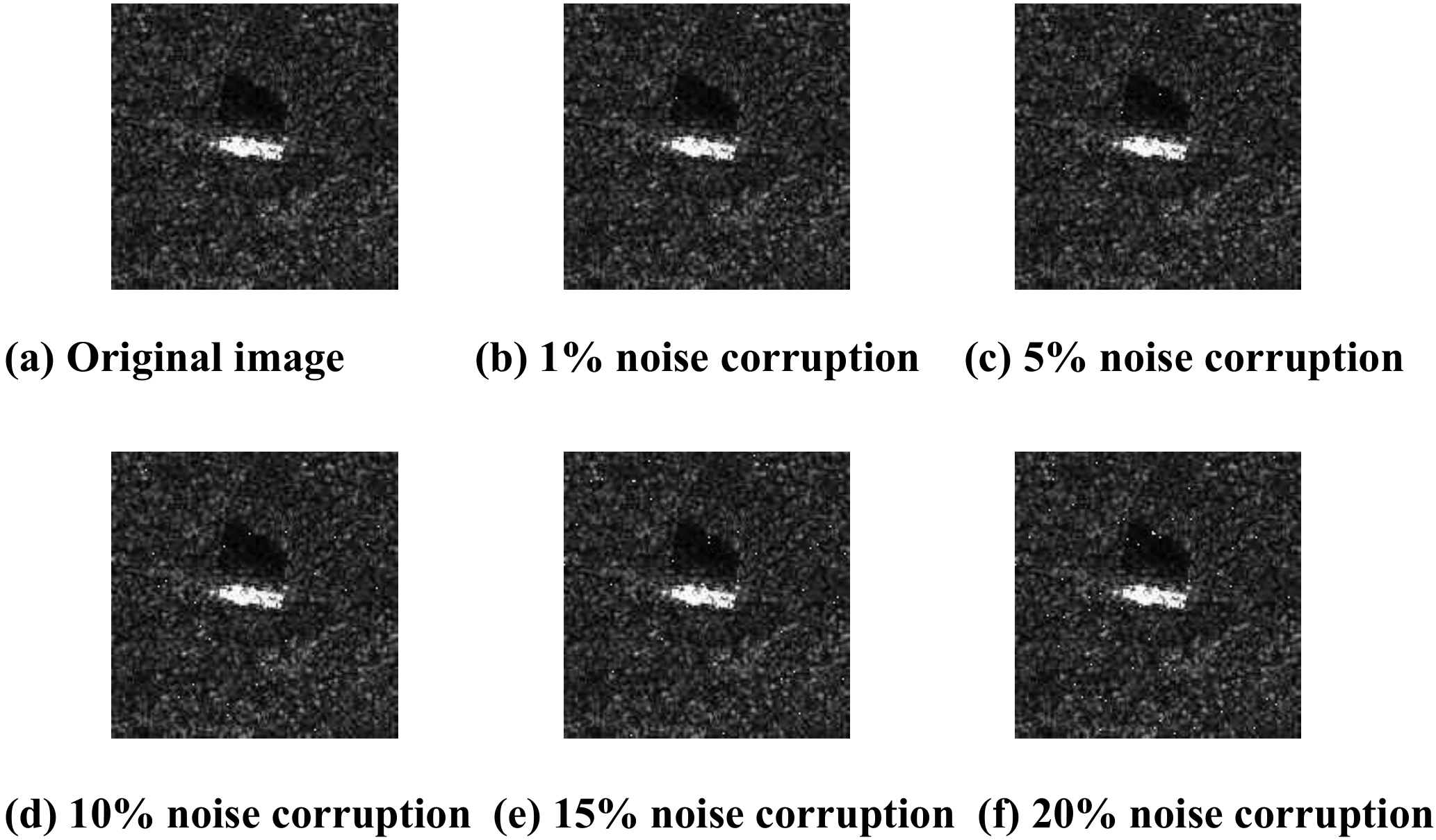}
	\caption{Illustration of random noise corruption. (a) is the original image. (b)-(f) are images respectively with 1\%, 5\%, 10\%, 15\% and 20\% noise corruption.}
	\label{noise}
\end{figure}

\begin{figure}[]
	\centering
	\includegraphics[width=3.5in]{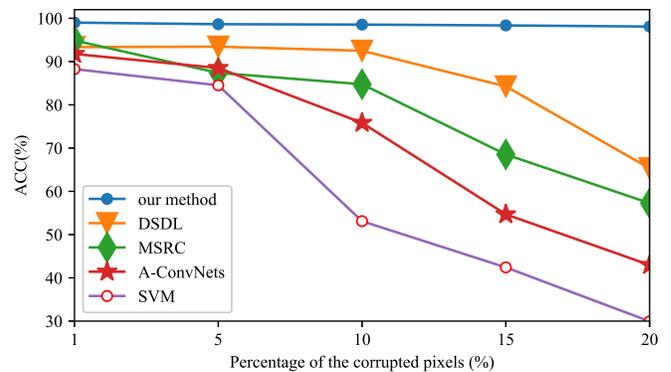}
	\caption{Average accuracy curves under different algorithms with different percentages of noise corruption.}
	\label{noise_compare}
\end{figure}

In order to test the anti-noise performance of our model, we randomly select a certain proportion of pixels in the test dataset and replace them with samples generated from a uniform distribution, which is shown in Fig \ref{noise}. This noise simulation method is consistent with the approach in \cite{chen2016target, NoneSparse} and the variance of noise can be easily obtained by many method \cite{ulfarsson2008dimension, chu2019eigen}. With the network previously trained by the ten-class SAR dataset, we feed these test images into our model. The anti-noise performance is shown in Fig \ref{noise_compare}, in which we compared the proposed method with four competing methods: SVM, A-ConvNets, MSRC, DSDL. The result shows that with noise proportion added to 20\%, the accuracy of our method is still beyond 98\%, while the other four methods have a comparatively significant drop.

\subsection{Detection and Recognition on Large-scene SAR Images}
The large-scene SAR images used in this paper were also collected under an X-band SAR sensor, in a 1-ft resolution spotlight mode, with a high resolution of $0.3\times0.3$ in both range and azimuth, which is the same as the small target chips. Therefore, it is reasonable to embed these targets from the $128\times128$ image chips into the large-scene SAR images.

When conducting experiments on complex large-scene SAR images, the evaluation indicators we use are accuracy (ACC), False Negative Rate (FNR), and False Positive Rate (FPR). FNR indicates how many targets we have missed or not correctly detected during detection, and FPR demonstrates the probability that we recognize clutters in the background as the true targets. FPR is defined in Eq. (\ref{FPR}):
\begin{equation}
\label{FPR}
FPR = \frac{{FP}}{{TN + FP}}
\end{equation}
\subsubsection{Experiments on target segmentation and synthesis method}
\begin{figure}[]
	\centering
	\includegraphics[width=3.5in]{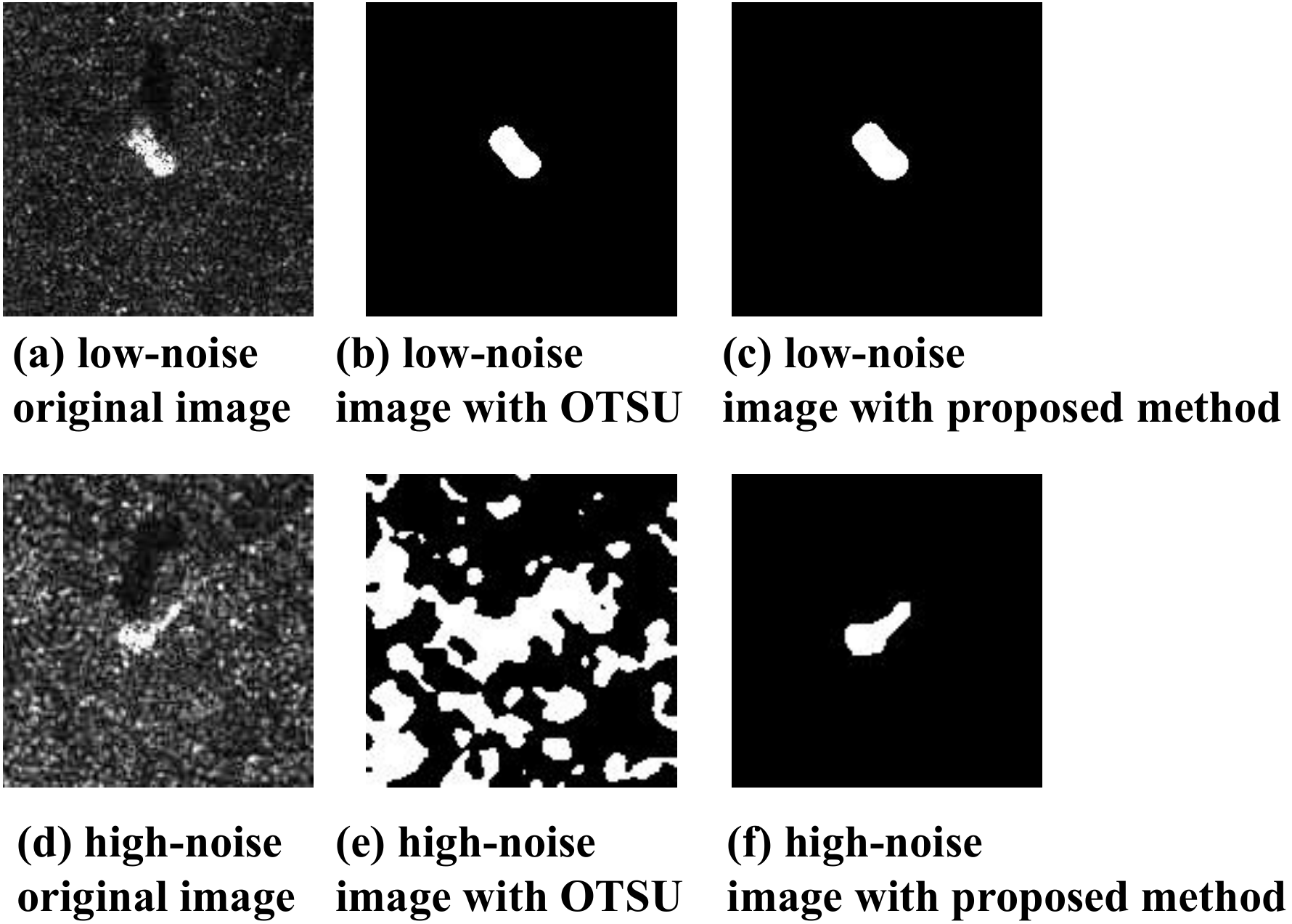}
	\caption{Comparison of OTSU and proposed method on low-noise SAR image and high-noise SAR image for object segmentation without shadow. (a) and (d) are respectively low-noise image and high-noise image; (b) and (e) are results of OTSU method; (c) and (f) are results of proposed method.}
	\label{highlowcom}
\end{figure}

For the purpose of selecting the appropriate threshold to segment the object without shadow from small target chips, we compare our thresholding method with the most popular adaptive thresholding method OTSU. These two methods are applied to the MSTAR ten-class target chips and the result is summarized in Table \ref{threshold_compare}.
The accuracy (ACC) denotes the percentage of the targets which can be correctly segmented from the target chips. Our proposed thresholding method leads to a better performance than the adaptive thresholding method OTSU. As to target D7, T62, ZSU234, BMP2, BTR70 and T72, which have comparatively lower noise, our proposed method does not have an obvious advantage in accuracy. However, when encountering with higher background noises,  such as the target 2S1, BRDM2, BTR60 and ZIL131, the accuracy using OTSU drops significantly, especially in BRDM2 and BTR60, where OTSU cannot segment over a half of the total images. The segmentation result on low and high noise images based on these two methods is shown in Fig. \ref{highlowcom}. Besides, the average accuracy of our method reaches 96.28\%, 16.33\% higher than OTSU. Therefore, the proposed thresholding method has a more stable performance as to SAR target segmentation.
\begin{table*}[]
	\centering
	\caption{Thresholding method comparison on MSTAR ten-class target chips.}
	\resizebox{\textwidth}{!}
	{
		\begin{tabular}{cc|ccccccccccc}
			\toprule[2pt]
			&                         & \textbf{2S1} & \textbf{BRDM2} & \textbf{BTR60} & \textbf{D7} & \textbf{T62} & \textbf{ZIL131} & \textbf{ZSU234} & \textbf{BMP2} & \textbf{BTR70} & \textbf{T72} & \textbf{Average} \\ \hline
			\multicolumn{2}{c|}{\textbf{Number}}                                              & 274          & 274            & 195            & 274         & 273          & 274             & 299             & 233           & 233            & 232          & -                \\ \cline{1-2}
			\multicolumn{1}{c|}{\multirow{2}{*}{\textbf{Our method}}} & \textbf{Threshold(\%)} & 92           & 88             & 90             & 92          & 90           & 90              & 95              & 95            & 95             & 95           & -                \\ \cline{2-2}
			\multicolumn{1}{c|}{}                                   & \textbf{Acc(\%)}        & 93.43        & 87.23          & 86.67          & 99.27       & 99.65        & 95.11           & 100             & 100           & 100            & 100          & 96.28            \\ \cline{1-2}
			\multicolumn{1}{c|}{\textbf{OTSU}}                      & \textbf{Acc(\%)}        & 53.65        & 32.85          & 43.08          & 98.91       & 88.64        & 79.06           & 100             & 100           & 100            & 100          & 79.95            \\ \bottomrule[2pt]
			\label{threshold_compare}
		\end{tabular}
	}
\end{table*}

\begin{figure}[]
	\centering
	\includegraphics[width=3.5in]{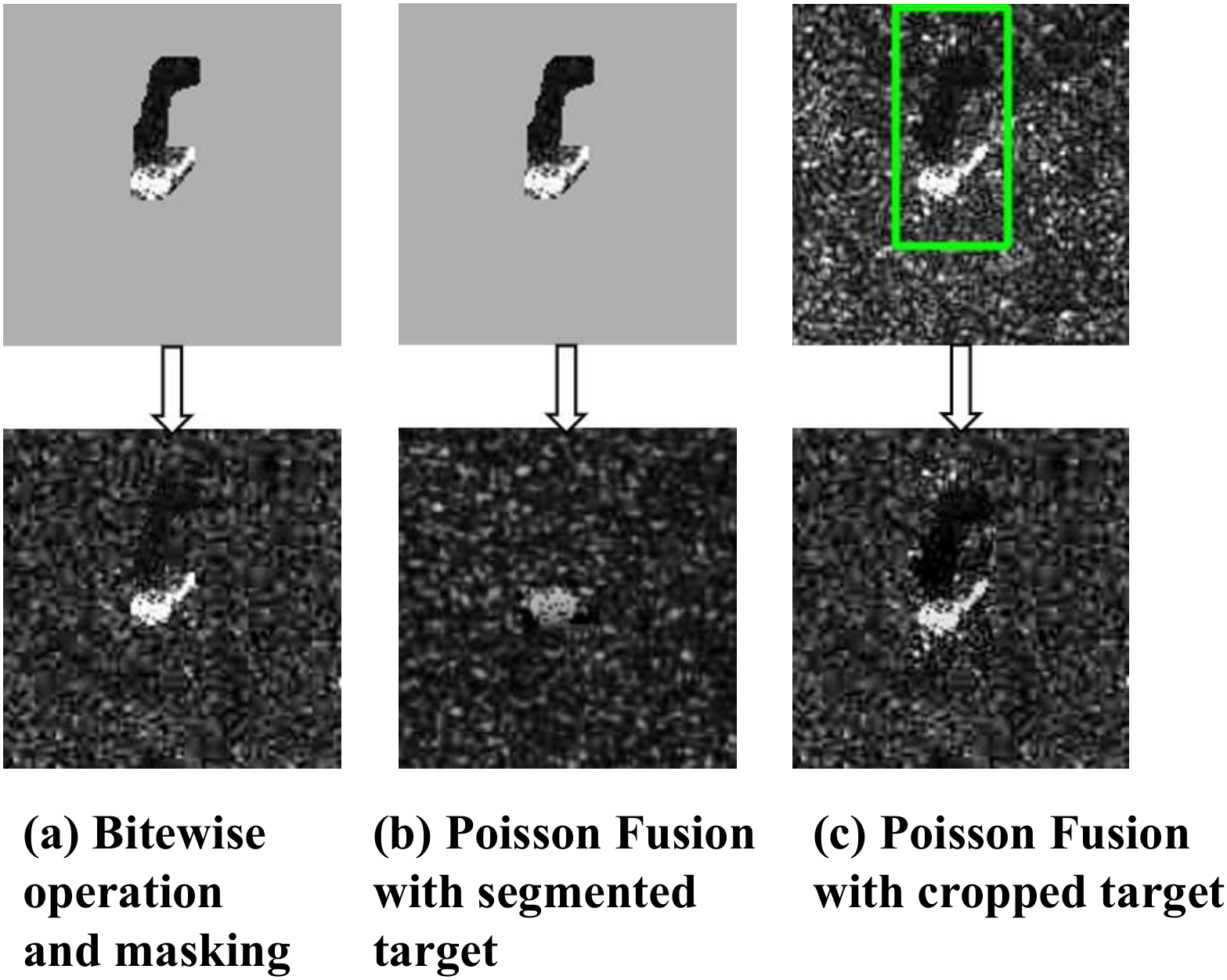}
	\caption{Comparison of different target synthesis methods. (a) is the result of the method that we finally choose: bitewise operation and masking; (b) is the result of Poisson Fusion with segmented target; (c) is the result of Poisson Fusion with cropped target.}
	\label{synthesistarget}
\end{figure}

As shown in Fig. \ref{synthesistarget}, the method bitwise operation and masking has better synthesis performance, since the object and its shadow are infused naturally into the background. The Poisson image fusion with segmented target softens the edges to a large extent so that it is hard to recognize the infused target. The third method which is to cut a small piece of the original SAR image chip and apply Poisson image fusion, solves the problem of over-softening, but the edge of the original image is still recognizable and it is hard to match the various background of target chips to the same large-scene image. In summary, the first method exhibits better generalization performance and is more suitable for the SAR target synthesis task.
\begin{figure*}[]
	\centering
	\includegraphics[width=1\textwidth]{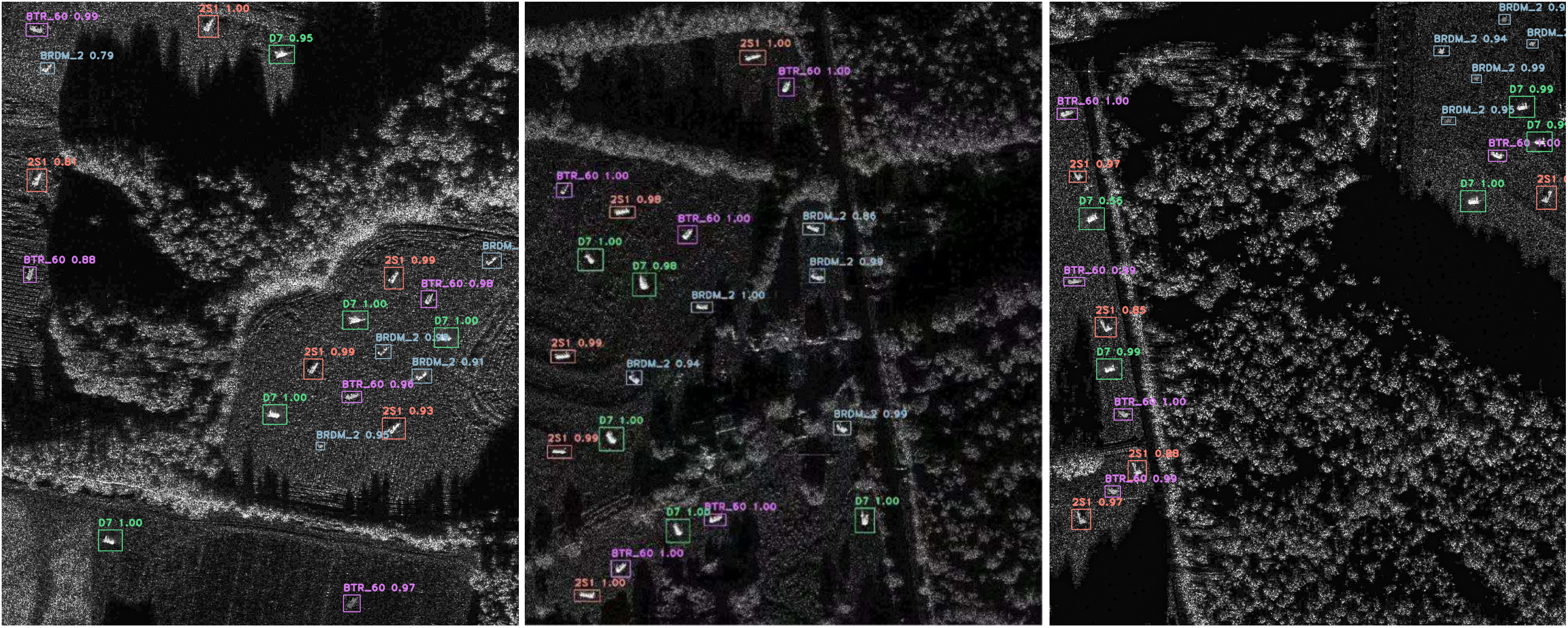}
	\caption{Detection result on three complex large-scene SAR images.}
	\label{large_eg}
\end{figure*}

\subsubsection{Detection and recognition on complex large-scene SAR images}

\begin{figure*}[]
	\centering
	\includegraphics[width=1\textwidth]{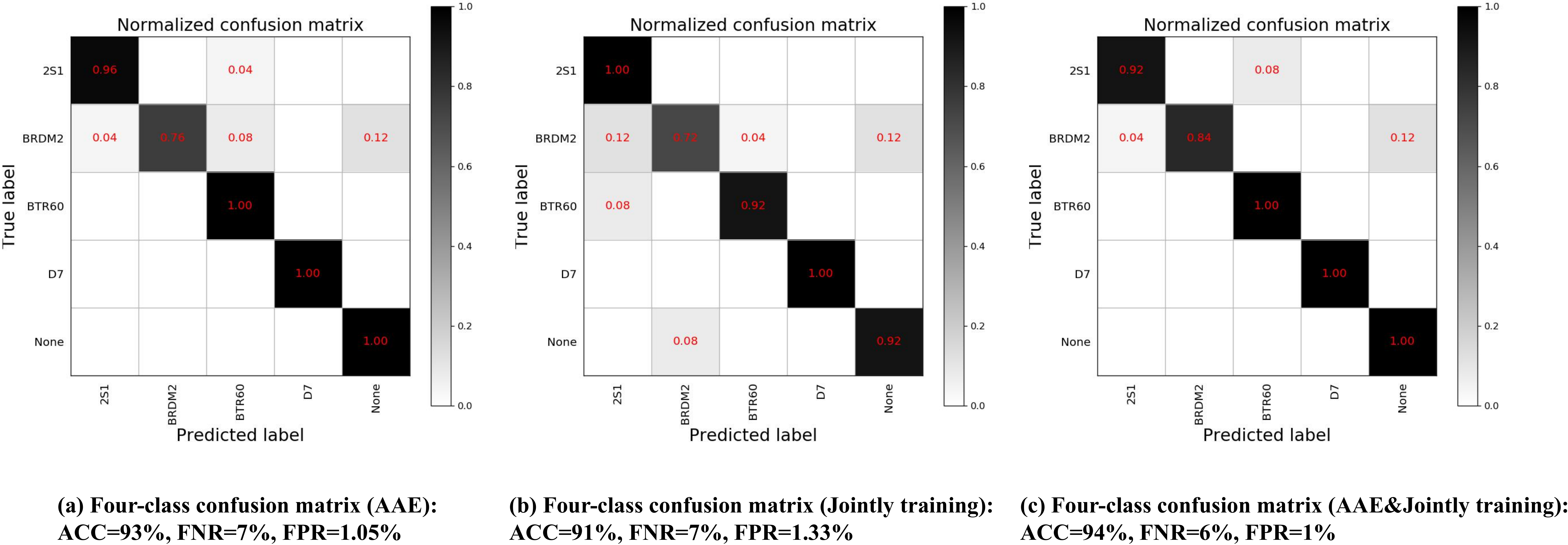}
	\caption{Normalized confusion matrix of large-scene SAR images. (a) is the result with AAE data augmentation method; (b) is the result with jointly training strategy; (c) is the result with AAE data augmentation method and jointly training strategy.}
	\label{matrix}
\end{figure*}

Four MSTAR targets including 2S1, BRDM2, BTR60 and D7 are chosen to be synthesized on the complex large-scene background. There are five same category targets on one large-scene image, which means 20 targets in each image. 
In the training process, 35 synthesized large-scene images are randomly chosen and divided into four different sizes ($128 \times 128$, $256 \times 256$, $512 \times 512$, $1024 \times 1024$), and 5 synthesized images are divided for testing. The final size $1024 \times 1024$ is chosen to conduct fast sliding on the testing images, which has the highest detection accuracy rate in the validation part since it is of better robustness with more complex background information. After fast sliding, the total number of training images is 4338, which includes 1091 large-scene images and 3247 expanded small target chips. The number of testing images is 150. The detection results is shown in Fig. \ref{large_eg}. Fig. \ref{matrix} shows the normalized confusion matrix, which reflects four-class targets detection outcome. The accuracy rate raises from 93\% to 94\% after jointly training, and from 91\% to 94\% after data augmentation by AAE. In the meantime, by combining jointly training strategy and AAE method, FNR decreases by 1\%, and FPR drops from 1.33\% to 1\%. 

Table \ref{largescene} shows the detection and recognition performance based on our proposed method and other widely used object detection methods. It is noted that the proposed method is a comprehensive detection framework including target segmentation and synthesis, data augmentation method AAE, automatic labeling, fast sliding and jointly training through YOLOv3 network. For large-scene image detection, it has been demonstrated by the experimental results that the proposed method has 23\% performance gain on accuracy when compared with the one directly applying YOLOv3 network. In addition, it is about 4.2 times faster than the well-established Faster R-CNN method, reaching a 5\% higher accuracy rate, and 3 times faster than SSD, which proves its superior real-time performance.

Thus, in respect of both effectiveness and efficiency, our method reaches 94\% on ACC and only cost 0.038s per image with 6\% FNR and 1\% FPR, which proves that it is a promising framework to deal with real-time detection and recognition on complex large-scene SAR images.
\begin{table}[]
	\centering
	\caption{Detection and recognition results on large-scene SAR images.}
	\resizebox{0.5\textwidth}{!}{
	\begin{tabular}{ccccc}
		\toprule[2pt]
		\textbf{Method}                & \textbf{ACC(\%)} & \textbf{FNR(\%)} & \textbf{FPR(\%)} & \textbf{Time cost for detection/seconds} \\ \hline
		\textbf{Our method}              & \textbf{94}               & \textbf{6}                & \textbf{1}                & \textbf{5.572}                                    \\
		\textup{YOLOv3 (applied with fast sliding)}                & 93               & 7               & 1.33                & 5.627                                    \\
		\textup{YOLOv3 (applied without fast sliding)}                & 71               & 15               & 2.73                & 2.506                                    \\
		\textup{Faster R-CNN (applied with fast sliding)} & 89               & 5                & 0                & 23.441                                  \\
		\textup{Faster R-CNN (applied without fast sliding)}          & 79               & 9                & 0                & 19.440                                   \\ 
		\textup{SSD (applied with fast sliding)} & 85               & 7                & 1.45                & 16.329                                  \\
		\textup{SSD (applied without fast sliding)}          & 73               & 13                &3.63                 & 9.267                                   \\ 
		\bottomrule[2pt]
		\label{largescene}
	\end{tabular}}
\end{table}

\begin{figure*}[]
	\centering
	\includegraphics[width=1\textwidth]{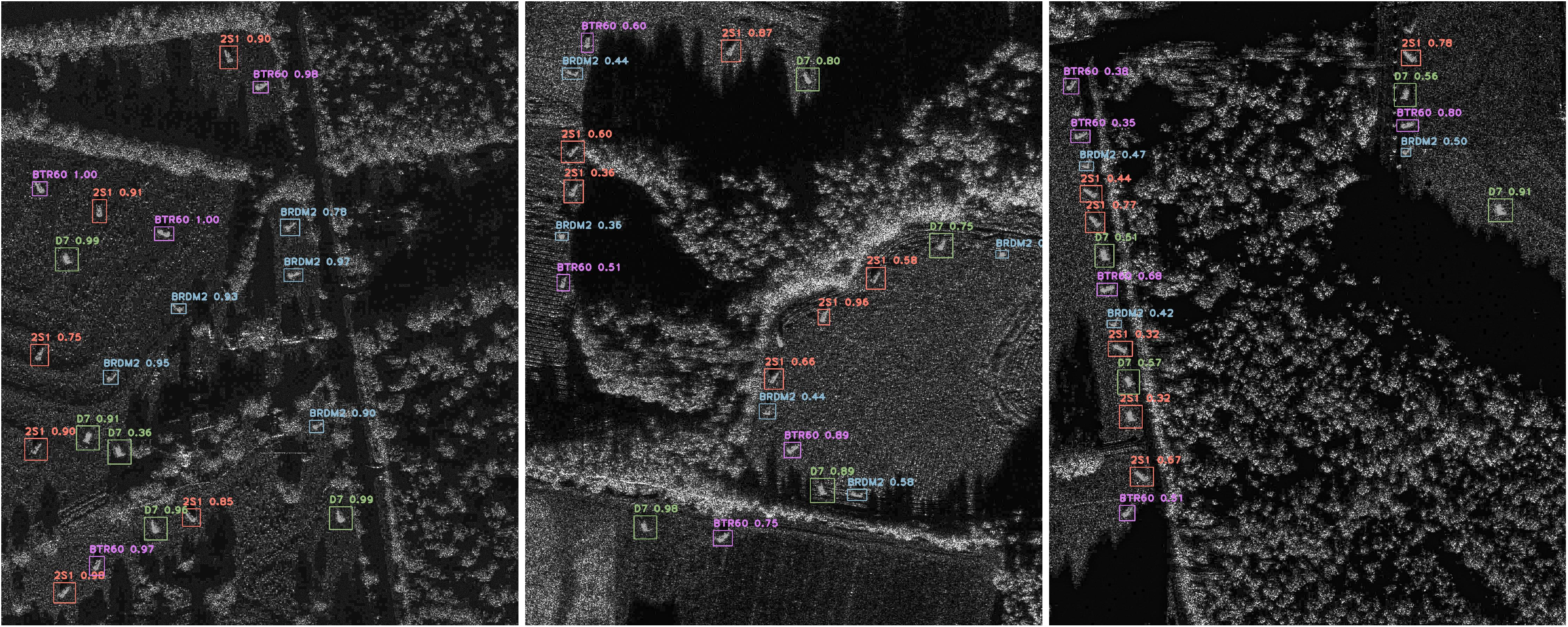}
	\caption{Detection results of synthesized images with darkened-targets and tricky position.}
	\label{target-darkened_result}
\end{figure*}

To further prove the robustness of our method, a more challenging experiment is conducted. We selected three large-scene images with the most complex background, and laid the targets alongside the trees, as is shown in the following Fig. \ref{target-darkened}. Besides, we darkened the targets by 40\%, so there is lower contrast between the metallic targets and their background. Then we use our original trained weight to perform detection, and the results are shown in Fig. \ref{target-darkened_result} and Table \ref{darkenresult}.

\begin{figure*}[]
	\centering
	\includegraphics[width=1\textwidth]{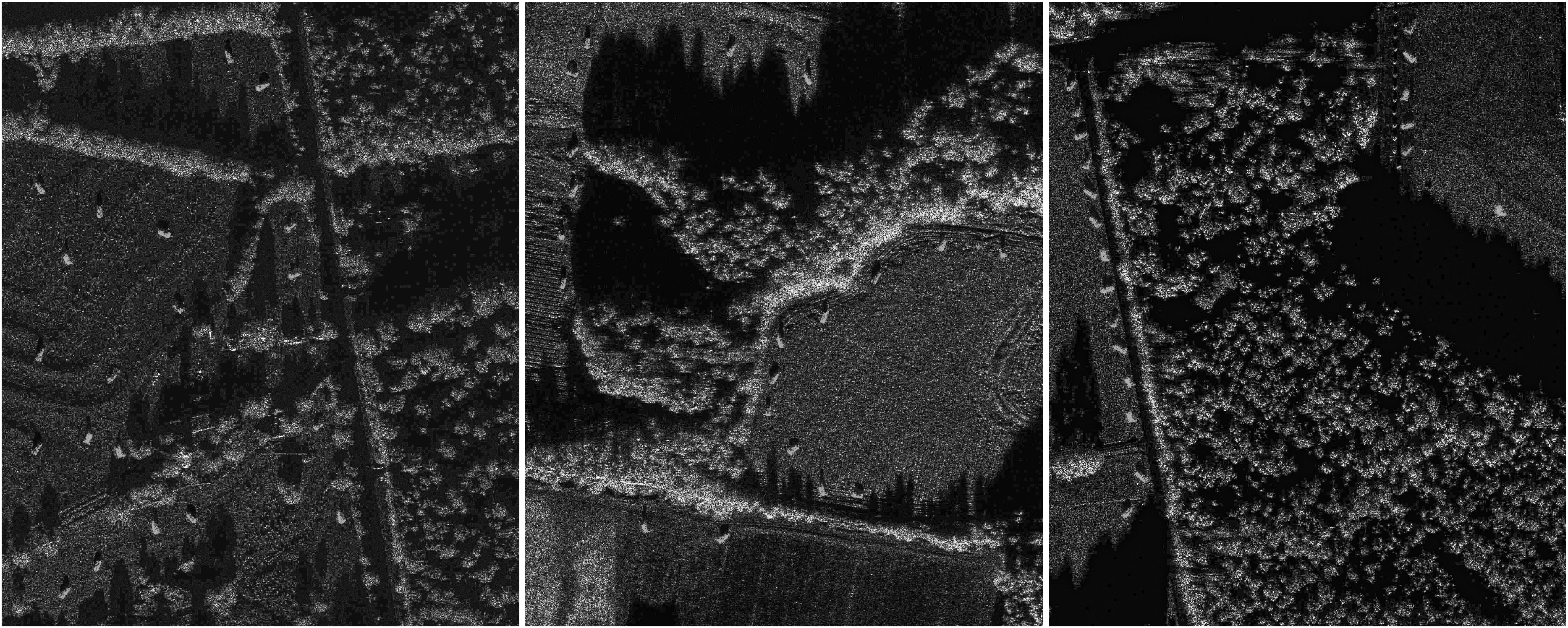}
	\caption{Synthesized images with darkened-targets and tricky position.}
	\label{target-darkened}
\end{figure*}

\begin{table}[]
\centering
\caption{Confusion matrix for four-class large-scene SAR image detection and recognition.}
\resizebox{0.5\textwidth}{!}{
\begin{tabular}{ccccccccc}
\toprule[2pt]

\textbf{class}   & \textbf{2S1} & \textbf{BRDM2} & \textbf{BTR60} & \textbf{D7} & \textbf{None} & \textbf{ACC(\%)} & \textbf{FNR(\%)} & \textbf{FPR(\%)} \\ \hline
\textbf{2S1}     & 15           & 0              & 0              & 0           & 0             & 100              & 0                & 0                \\
\textbf{BRDM2}   & 0            & 13             & 0              & 0           & 2             & 86.67            & 13.33            & 0                \\
\textbf{BTR60}   & 2            & 0              & 13             & 0           & 0             & 86.67            & 0                & 13.33            \\
\textbf{D7}      & 1            & 0              & 0              & 14          & 0             & 93.33            & 0                & 7.14             \\
\textbf{Average} &              &                &                &             &               & \textbf{91.67}   & \textbf{3.33}    & \textbf{5.12}    \\ 
\bottomrule[2pt]
\label{darkenresult}
\end{tabular}
}
\end{table}

It is seen from Table \ref{darkenresult} that low contrast leads to targets been treated as background, since being close to the tree brings much speckle noise interruption and these trees have nearly equal brightness with the targets. Besides, BTR60 and D7 have been recognized as 2S1. The reason may be that 2S1 target is darker as a whole, so the darkening process makes it hard to distinguish some objects from 2S1. However, the proposed method's average accuracy remains above 91\% even under such a tricky condition, proving the effectiveness and robustness of the proposed method.

\section{Discussions and Conclusions}\label{4}
\subsection{Discussions}
In this subsection, we will discuss several experimental results. \textbf{Firstly}, in data augmentation experiments, we found that the generated images by GANs are of far less variety when training epochs are under 200, and it costs much time and space to generate images when the size reaches $128 \times 128$. On the contrary, the images generated by the AAE framework are of high quality and rich diversity. Besides, the training process of the AAE framework can be completed in an efficient manner (the stable result can be obtained within 200 epochs, costing less than 3 seconds). What's more, the FID score of AAE is the lowest compared with GAN-based methods, which proves that the generated images are of richer diversity, thus we choose AAE as our data augmentation method. 

\textbf{Secondly}, we conduct an experiment of target recognition on ten-class MSTAR dataset. From Table \ref{with_aae}, we can see that some targets in class BRDM, BTR60, D7, T62 and BMP2 are recognized as ZSU234, which leads to the decrease of final detection accuracy. The reason for that may be that the background of target ZSU234 is darker than other objects, thus the shadow around the targets may misguide the final judgment, as the recognition depends on the detected region. Besides, the FNR is largely caused by BTR60 since some targets in BTR60 have nearly the same pattern so the network treats the noise as the target, thus it is rather hard to tell them apart. But we can clearly see that the accuracy after data augmentation raises to nearly 99\%, and FNR drops from 2.025\% to 1.204\%, which proves the effectiveness of AAE data augmentation method on small target chips.

\textbf{Thirdly}, in the noise corruption experiment, we can see that our framework exhibits high noise immunity. As shown in Fig. \ref{noise_compare}, with noise proportion raises up to 20\%, the accuracy still remains above 98\%. Such superior performance can be explained by the fact that the proposed method is capable of telling the object from its background. As a result, the noise corruption in the background can be further learned as disturbance so that our method can significantly maintain the original object and recognize which category it belongs to.

\textbf{Finally}, the experiment conducted on large-scene images shows that the detection accuracy raises up by 1\% and FNR drops by 1\%, FPR drops by 0.33\% after we simultaneously train the expanded small SAR chips and sliced large-scene SAR images. This can be explained by noticing that the training process of small target chips can make the network learn more about the textural feature of SAR images since the target in small chips is comparatively a rather large object while in large-scene images is only a small object. Therefore, small chips act like a supplementary, assisting the recognition of SAR objects on large-scene images. Besides, the experiment on the more tricky dataset we provided proves that our network learns target textual feature rather than the pure edge information, therefore, it is much more robust. But the false positive cases still remain an intractable problem.
We suppose that the following ideas may be able to reduce the false positive cases. 1) Adopting more data augmentation methods on the objects which are easily detected incorrect, enabling the model to learn more diverse target features and enhance the model's robustness; 2) Methods like hard example mining \cite{shrivastava2016training} and focal loss \cite{lin2017focal} will increase the weight of hard example in training process, which may be beneficial to minimize the false positive cases. While in some situations, however, the shadow may share the same feature with the targets. Under such condition, pre-training can be an effective method to solve this problem. For instance,  contrastive learning \cite{chen2020simple,momentum2020he}, targeting at learning an encoder that is able to map positive pairs to similar representations while push away those negative samples in the embedding space. In this way, the pre-trained model may have stronger generalization and feature extraction capability, thus effectively distinguishing target from its shadow. 

\subsection{Conclusions}

In this paper, an efficient and robust deep learning based target detection method has been proposed based on a novel customized learning representations and multi-scale features of SAR images method. The framework of AAE has been employed for advanced data augmentation which was confirmed by a high variety of the generated samples. An automatic labeling method has been proposed to avoid the labor-intensive manual labeling. By jointly training the neural network with the small target chips and large-scene images, the proposed integrated target detector has been proposed to realized multiple targets detection and recognition. The experimental results confirmed  our method reached competitive accuracy on complex large-scene SAR images with rapid speed. Besides, our method can obtain robust detection performance in terms of the different noise levels, even in the extreme case that the corrupted pixels reach 20\%.

It is noted that there are still some potential problems needed to be tackled in the future: 1) The detection accuracy varies among different categories, and some categories, such as BRDM2, are hard to recognize since their feature is similar to the background; 2) It was found that SAR targets have different rotation angles. Therefore, using rotated anchors to perform targets detection may enhance the final detection accuracy.

\ifCLASSOPTIONcaptionsoff
  \newpage
\fi



%
\bibliographystyle{IEEEtran}
\bibliography{mybibfile}

\begin{thebibliography}{10}
\providecommand{\url}[1]{#1}
\csname url@samestyle\endcsname
\providecommand{\newblock}{\relax}
\providecommand{\bibinfo}[2]{#2}
\providecommand{\BIBentrySTDinterwordspacing}{\spaceskip=0pt\relax}
\providecommand{\BIBentryALTinterwordstretchfactor}{4}
\providecommand{\BIBentryALTinterwordspacing}{\spaceskip=\fontdimen2\font plus
\BIBentryALTinterwordstretchfactor\fontdimen3\font minus
  \fontdimen4\font\relax}
\providecommand{\BIBforeignlanguage}[2]{{%
\expandafter\ifx\csname l@#1\endcsname\relax
\typeout{** WARNING: IEEEtran.bst: No hyphenation pattern has been}%
\typeout{** loaded for the language `#1'. Using the pattern for}%
\typeout{** the default language instead.}%
\else
\language=\csname l@#1\endcsname
\fi
#2}}
\providecommand{\BIBdecl}{\relax}
\BIBdecl

\bibitem{wang2016robust}
H.~Wang, Y.~Cai, G.~Fu, and S.~Wang, ``Robust automatic target recognition
  algorithm for large-scene sar images and its adaptability analysis on
  speckle,'' \emph{Science Program.}, vol. 2016, 2016.

\bibitem{liu2018sar}
M.~Liu, S.~Chen, X.~Wang, F.~Lu, M.~Xing, and J.~Wu, ``Sar target configuration
  recognition via discriminative statistical dictionary learning,'' \emph{IEEE
  J. Sel. Top. Appl. Earth Observ. Remote Sens.}, vol.~11, no.~11, pp.
  4218--4229, 2018.

\bibitem{zhang2018fast}
Y.~Zhang, Y.~Song, Y.~Wang, and H.~Qu, ``A fast training method for sar large
  scale samples based on cnn for targets recognition,'' in \emph{11th Int.
  Congr. on Image and Signal Process., Biomed. Eng. and Inform.
  (CISP-BMEI)}.\hskip 1em plus 0.5em minus 0.4em\relax IEEE, 2018, pp. 1--5.

\bibitem{zhao2018adaptive}
Y.~Zhao and P.~Liu, ``Adaptive ship detection for single-look complex sar
  images based on svwie-noncircularity decomposition,'' \emph{Sensors},
  vol.~18, no.~10, p. 3293, 2018.

\bibitem{8533426}
Q.~{Lu}, Y.~{Gao}, P.~{Huang}, and X.~{Liu}, ``Range- and aperture-dependent
  motion compensation based on precise frequency division and chirp scaling for
  synthetic aperture radar,'' \emph{IEEE Sensors Journal}, vol.~19, no.~4, pp.
  1435--1442, 2019.

\bibitem{dudgeon1993overview}
B.~{Bhanu}, ``Automatic target recognition: State of the art survey,''
  \emph{IEEE Trans. Aerosp. Electron. Syst.}, vol. AES-22, no.~4, pp. 364--379,
  1986.

\bibitem{morgan2015deep}
D.~A. Morgan, ``Deep convolutional neural networks for atr from sar imagery,''
  in \emph{Algorithms for Synthetic Aperture Radar Imagery XXII}, vol.
  9475.\hskip 1em plus 0.5em minus 0.4em\relax Int. Soc. for Opt. and
  Photonics, 2015, p. 94750F.

\bibitem{article}
E.~Keydel, S.~Lee, and J.~Moore, ``Mstar extended operating conditions - a
  tutorial,'' \emph{Proc. of SPIE - The Int. Soc. for Opt. Eng.}, 06 1996.

\bibitem{cui2019d}
Z.~Cui, C.~Tang, Z.~Cao, and N.~Liu, ``D-atr for sar images based on deep
  neural networks,'' \emph{Remote Sens.}, vol.~11, no.~8, p. 906, 2019.

\bibitem{wen2018survey}
F.~Wen, L.~Chu, P.~Liu, and R.~C. Qiu, ``A survey on nonconvex
  regularization-based sparse and low-rank recovery in signal processing,
  statistics, and machine learning,'' \emph{IEEE Access}, vol.~6, pp.
  69\,883--69\,906, 2018.

\bibitem{1202937}
B.~{Krishnapuram}, J.~{Sichina}, and L.~{Carin}, ``Physics-based detection of
  targets in sar imagery using support vector machines,'' \emph{IEEE Sensors
  Journal}, vol.~3, no.~2, pp. 147--157, 2003.

\bibitem{inproceedings}
Y.~Wang, Y.~Zhang, H.~Qu, and Q.~Tian, ``Target detection and recognition based
  on convolutional neural network for sar image,'' in \emph{11th Int. Congr. on
  Image and Signal Process., Biomed. Eng. and Inform.}, 2018, pp. 1--5.

\bibitem{ding2016convolutional}
J.~Ding, B.~Chen, H.~Liu, and M.~Huang, ``Convolutional neural network with
  data augmentation for sar target recognition,'' \emph{IEEE Geosci. Remote
  Sens. Lett.}, vol.~13, no.~3, pp. 364--368, 2016.

\bibitem{chen2016target}
S.~Chen, H.~Wang, F.~Xu, and Y.-Q. Jin, ``Target classification using the deep
  convolutional networks for sar images,'' \emph{IEEE Trans. Geosci. Remote
  Sensing}, vol.~54, no.~8, pp. 4806--4817, 2016.

\bibitem{goodfellow2014generative}
I.~Goodfellow, J.~Pouget-Abadie, M.~Mirza, B.~Xu, D.~Warde-Farley, S.~Ozair,
  A.~Courville, and Y.~Bengio, ``Generative adversarial nets,'' in \emph{Adv.
  in Neural Inf. Process. Syst.}, 2014, pp. 2672--2680.

\bibitem{he2019parallel}
C.~He, D.~Xiong, Q.~Zhang, and M.~Liao, ``Parallel connected generative
  adversarial network with quadratic operation for sar image generation and
  application for classification,'' \emph{Sensors}, vol.~19, no.~4, p. 871,
  2019.

\bibitem{ShiAutomatic}
X.~Shi, F.~Zhou, S.~Yang, Z.~Zhang, and T.~Su, ``Automatic target recognition
  for synthetic aperture radar images based on super-resolution generative
  adversarial network and deep convolutional neural network,'' \emph{Remote
  Sens.}, vol.~11, no.~2, p. 135, 2019.

\bibitem{makhzani2015adversarial}
A.~Makhzani, J.~Shlens, N.~Jaitly, I.~Goodfellow, and B.~Frey, ``Adversarial
  autoencoders,'' \emph{arXiv preprint arXiv:1511.05644}, 2015.

\bibitem{li2017prediction}
H.~Li and S.~Misra, ``Prediction of subsurface nmr t2 distributions in a shale
  petroleum system using variational autoencoder-based neural networks,''
  \emph{IEEE Geosci. Remote Sens. Lett.}, vol.~14, no.~12, pp. 2395--2397,
  2017.

\bibitem{xiao2020deep}
F.~Xiao, L.~Pei, L.~Chu, D.~Zou, W.~Yu, Y.~Zhu, and T.~Li, ``A deep learning
  method for complex human activity recognition using virtual wearable
  sensors,'' \emph{arXiv preprint arXiv:2003.01874}, 2020.

\bibitem{girshick2014rich}
R.~Girshick, J.~Donahue, T.~Darrell, and J.~Malik, ``Rich feature hierarchies
  for accurate object detection and semantic segmentation,'' in \emph{Proc. of
  the IEEE Conf. on Comput. Vis. and Pattern Recognit.}, 2014, pp. 580--587.

\bibitem{girshick2015fast}
R.~Girshick, ``Fast r-cnn,'' in \emph{Proc. of the IEEE Int. Conf. on Comput.
  Vis.}, 2015, pp. 1440--1448.

\bibitem{ren2015faster}
S.~Ren, K.~He, R.~Girshick, and J.~Sun, ``Faster r-cnn: Towards real-time
  object detection with region proposal networks,'' in \emph{Adv. in Neural
  Inf. Process. Syst.}, 2015, pp. 91--99.

\bibitem{liu2016ssd}
W.~Liu, D.~Anguelov, D.~Erhan, C.~Szegedy, S.~Reed, C.-Y. Fu, and A.~C. Berg,
  ``Ssd: Single shot multibox detector,'' in \emph{Eur. Conf. on Comput.
  Vis.}\hskip 1em plus 0.5em minus 0.4em\relax Springer, 2016, pp. 21--37.

\bibitem{redmon2016you}
J.~Redmon, S.~Divvala, R.~Girshick, and A.~Farhadi, ``You only look once:
  Unified, real-time object detection,'' in \emph{Proc. of the IEEE Conf. on
  Comput. Vis. and Pattern Recognit.}, 2016, pp. 779--788.

\bibitem{redmon2017yolo9000}
J.~Redmon and A.~Farhadi, ``Yolo9000: better, faster, stronger,'' in
  \emph{Proc. of the IEEE Conf. on Comput. Vis. and Pattern Recognit.}, 2017,
  pp. 7263--7271.

\bibitem{redmon2018yolov3}
------, ``Yolov3: An incremental improvement,'' \emph{arXiv preprint
  arXiv:1804.02767}, 2018.

\bibitem{dong2019end}
M.~Dong, Y.~Cui, X.~Jing, X.~Liu, and J.~Li, ``End-to-end target detection and
  classification with data augmentation in sar images,'' in \emph{2019 IEEE
  Int. Conf. on Comput. Electromagn. (ICCEM)}.\hskip 1em plus 0.5em minus
  0.4em\relax IEEE, 2019, pp. 1--3.

\bibitem{huang2015densebox}
L.~Huang, Y.~Yang, Y.~Deng, and Y.~Yu, ``Densebox: Unifying landmark
  localization with end to end object detection,'' \emph{arXiv preprint
  arXiv:1509.04874}, 2015.

\bibitem{Tian2018Robust}
L.~Tian, M.~Li, Y.~Hao, J.~Liu, G.~Zhang, and Y.~Q. Chen, ``Robust 3-d human
  detection in complex environments with a depth camera,'' \emph{IEEE Trans.
  Multimedia}, vol.~20, no.~9, pp. 2249--2261, 2018.

\bibitem{ivavsic2019human}
M.~Iva{\v{s}}i{\'c}-Kos, M.~Kri{\v{s}}to, and M.~Pobar, ``Human detection in
  thermal imaging using yolo,'' in \emph{Proc. of the 5th Int. Conf. on Comput.
  and Technol. Appl.}, 2019, pp. 20--24.

\bibitem{chang2010change}
N.~Chang, M.~Han, W.~Yao, L.~Chen, and S.~Xu, ``Change detection of land use
  and land cover in an urban region with spot-5 images and partial lanczos
  extreme learning machine,'' \emph{J. Appl. Remote Sens.}, vol.~4, no.~1, p.
  043551, 2010.

\bibitem{xu2019retrieval}
M.~Xu, H.~Xiang, H.~Yun, X.~Ni, W.~Chen, and C.~Cao, ``Retrieval of forest
  canopy height jointly using airborne lidar and alos palsar data,'' \emph{J.
  Appl. Remote Sens.}, vol.~14, no.~02, p. 022203, 2019.

\bibitem{wang2018sar}
Z.~Wang, L.~Du, J.~Mao, B.~Liu, and D.~Yang, ``Sar target detection based on
  ssd with data augmentation and transfer learning,'' \emph{IEEE Geosci. Remote
  Sens. Lett.}, vol.~16, no.~1, pp. 150--154, 2018.

\bibitem{Dowson1982The}
D.~C. Dowson and B.~V. Landau, ``The frechet distance between multivariate
  normal distributions,'' \emph{J. Multivar. Anal.}, vol.~12, no.~3, pp.
  450--455, 1982.

\bibitem{O2001SAR}
J.~A. O'Sullivan, M.~DeVore, V.~Kedia, and M.~I. Miller, ``Sar atr performance
  using a conditionally gaussian model,'' \emph{IEEE Trans. Aerosp. Electron.
  Syst.}, vol.~37, no.~1, pp. 91--108, 2001.

\bibitem{Zhao2001Support}
Q.~Zhao and J.~C. Principe, ``Support vector machines for sar automatic target
  recognition,'' \emph{IEEE Trans. Aerosp. Electron. Syst.}, vol.~37, no.~2,
  pp. 643--654, 2001.

\bibitem{Sun2007Adaptive}
Y.~Sun, Z.~Liu, S.~Todorovic, and J.~Li, ``Adaptive boosting for sar automatic
  target recognition,'' \emph{IEEE Trans. Aerosp. Electron. Syst.}, vol.~43,
  no.~1, pp. 112--125, 2007.

\bibitem{NoneSparse}
K.~G. Dong~G, Wang~N, ``Sparse representation of monogenic signal: With
  application to target recognition in sar images,'' \emph{IEEE Signal Process.
  Lett.}, vol.~21, no.~8, pp. 952--956.

\bibitem{DongClassification}
G.~Dong and G.~Kuang, ``Classification on the monogenic scale space:
  Application to target recognition in sar image,'' \emph{IEEE Trans. Image
  Process.}, vol.~24, no.~8, pp. 2527--2539.

\bibitem{Dong2017SAR}
G.~Dong, G.~Kuang, W.~Na, L.~Zhao, and J.~Lu, ``Sar target recognition via
  joint sparse representation of monogenic signal,'' \emph{IEEE J. Sel. Top.
  Appl. Earth Observ. Remote Sens.}, vol.~8, no.~7, pp. 3316--3328, 2017.

\bibitem{song2016sar}
S.~Song, B.~Xu, and J.~Yang, ``Sar target recognition via supervised
  discriminative dictionary learning and sparse representation of the sar-hog
  feature,'' \emph{Remote Sens.}, vol.~8, no.~8, p. 683, 2016.

\bibitem{Sun2016SAR}
Y.~Sun, D.~Lan, W.~Yan, Y.~Wang, and H.~Jing, ``Sar automatic target
  recognition based on dictionary learning and joint dynamic sparse
  representation,'' \emph{IEEE Geosci. Remote Sens. Lett.}, vol.~PP, no.~99,
  pp. 1--5, 2016.

\bibitem{huang2017transfer}
Z.~Huang, Z.~Pan, and B.~Lei, ``Transfer learning with deep convolutional
  neural network for sar target classification with limited labeled data,''
  \emph{Remote Sens.}, vol.~9, no.~9, p. 907, 2017.

\bibitem{Gu2018Fast}
Y.~Gu and Y.~Xu, ``Fast sar target recognition based on random convolution
  features and ensemble extreme learning machines,'' \emph{Opto-Electron.
  Eng.}, vol.~45, no.~1, 2018.

\bibitem{ulfarsson2008dimension}
M.~O. Ulfarsson and V.~Solo, ``Dimension estimation in noisy pca with sure and
  random matrix theory,'' \emph{IEEE transactions on signal processing},
  vol.~56, no.~12, pp. 5804--5816, 2008.

\bibitem{chu2019eigen}
L.~Chu, F.~Wen, and R.~C. Qiu, ``Eigen-inference precoding for coarsely
  quantized massive mu-mimo system with imperfect csi,'' \emph{IEEE
  Transactions on Vehicular Technology}, vol.~68, no.~9, pp. 8729--8743, 2019.

\bibitem{shrivastava2016training}
A.~Shrivastava, A.~Gupta, and R.~Girshick, ``Training region-based object
  detectors with online hard example mining,'' in \emph{IEEE Conf. on Comput.
  Vis. and Pattern Recognit.}, 2016, pp. 761--769.

\bibitem{lin2017focal}
Lin, Y.~T., P.~Goyal, R.~Girshick, and et~al, ``Focal loss for dense object
  detection,'' in \emph{IEEE Int. Conf. on Comput. Vis.}, 2017, pp. 2999--3007.

\bibitem{chen2020simple}
T.~Chen, S.~Kornblith, M.~Norouzi, and et~al, ``A simple framework for
  contrastive learning of visual representations,'' in \emph{Int. Conf. on
  Mach. Learn.}, 2020, pp. 1597--1607.

\bibitem{momentum2020he}
K.~He, H.~Fan, Y.~Wu, and et~al, ``Momentum contrast for unsupervised visual
  representation learning,'' in \emph{Proc. of the IEEE/CVF Conf. on Comput.
  Vis. and Pattern Recognit.}, 2020, pp. 9729--9738.

\end{thebibliography}




%

\begin{IEEEbiography}{Michael Shell}
Biography text here.
\end{IEEEbiography}

\begin{IEEEbiographynophoto}{John Doe}
Biography text here.
\end{IEEEbiographynophoto}


\begin{IEEEbiographynophoto}{Jane Doe}
Biography text here.
\end{IEEEbiographynophoto}




\end{document}